\documentclass[10pt,conference]{IEEEtran}
\IEEEoverridecommandlockouts
\let\labelindent\relax
\usepackage{amssymb}
\usepackage{cite}
\usepackage{pifont}
\usepackage{graphicx}
\usepackage{subfigure}
\usepackage{amsmath}
\usepackage{balance}
\usepackage{array}
\graphicspath{{figures/}}
\usepackage{subcaption}
\usepackage{amsmath}
\usepackage{bigstrut,multirow,rotating,booktabs}
\usepackage[authormarkup=none]{changes}
\usepackage{multirow}
\usepackage{algorithm}
\usepackage{algorithmicx}
\usepackage[noend]{algpseudocode}
\usepackage{framed}
\usepackage{tikz}
\usepackage{hyperref}
\usepackage[framemethod=TikZ]{mdframed} 
\usepackage{listings}
\usepackage{xcolor}
\usepackage{listings}
\usepackage{enumitem}
\usepackage{tcolorbox}
\usepackage{orcidlink}
\usepackage{xurl}

\makeatletter

\newcommand{\Rmnum}[1]{\expandafter\@slowromancap\romannumeral #1@}
\makeatother

\definecolor{royalblue}{RGB}{65,105,225}

\newcommand{\ga}[1]{\textcolor{blue}{#1}}

\definechangesauthor[color=red]{guoquan}

\definechangesauthor[color=red]{haomin}

\definechangesauthor[color=orange]{tian}

\definecolor{mygreen}{rgb}{0,0.6,0}  
\definecolor{mygray}{rgb}{0.5,0.5,0.5}  
\definecolor{mymauve}{rgb}{0.58,0,0.82}  

\definecolor{codegreen}{rgb}{0,0.6,0}
\definecolor{codegray}{rgb}{0.5,0.5,0.5}
\definecolor{codepurple}{rgb}{0.58,0,0.82}
\definecolor{backcolour}{rgb}{0.95,0.95,0.92}

\lstdefinestyle{mystyle}{
    backgroundcolor=\color{backcolour},   
    commentstyle=\color{codegreen},
    keywordstyle=\color{magenta},
    numberstyle=\tiny\color{codegray},
    stringstyle=\color{codepurple},
    basicstyle=\ttfamily\footnotesize,
    breakatwhitespace=false,         
    breaklines=true,                 
    captionpos=b,                    
    keepspaces=true,                 
    numbers=left,                    
    numbersep=5pt,                  
    showspaces=false,                
    showstringspaces=false,
    showtabs=false,                  
    tabsize=2
}

\graphicspath{{figures/}}

\usepackage{hyperref}
\hypersetup{
  colorlinks   = true,    
  urlcolor     = blue,    
  linkcolor    = blue,    
  citecolor    = blue      
}

\usepackage{xspace}

\begin{document}

\title{When Autonomous Vehicle Meets V2X Cooperative Perception: How Far Are We?}

\author{\IEEEauthorblockN{An Guo$^{1}\orcidlink{0009-0005-8661-6133}$,
Shuoxiao Zhang$^{1}\orcidlink{0009-0004-3023-5027}$, Enyi Tang$^{1}$\IEEEauthorrefmark{1}\orcidlink{0000-0001-9004-1292}\thanks{$^{*}$Enyi Tang and Zhenyu Chen are corresponding authors.}, Xinyu Gao$^{1}\orcidlink{0009-0004-7135-1833}$, Haomin Pang$^{2}\orcidlink{0009-0004-4237-7444}$, Haoxiang Tian$^{3}\orcidlink{0000-0001-9132-9319}$, \\Yanzhou Mu$^{1}\orcidlink{0000-0003-1816-2246}$, Wu Wen$^{2}\orcidlink{0000-0001-7727-5043}$, Chunrong Fang$^{1}\orcidlink{0000-0002-9930-7111}$ and Zhenyu Chen$^{4}$\IEEEauthorrefmark{1}\orcidlink{0000-0002-9592-7022}}
\IEEEauthorblockA{$^{1}$State Key Laboratory for Novel Software Technology, Nanjing University, China\\\{guoan218, sx\_zhang, xinyugao, muyanzhou\}@smail.nju.edu.cn, \{eytang, fangchunrong\}@nju.edu.cn
}
\IEEEauthorblockA{$^{2}$Guangzhou University, China, panghaomin@e.gzhu.edu.cn, wenwu@gzhu.edu.cn
}
\IEEEauthorblockA{$^{3}$Nanyang Technological University, Singapore, haoxiang.tian@ntu.edu.sg 
}
\IEEEauthorblockA{$^{4}$Shenzhen Research Institute of Nanjing University, China, zychen@nju.edu.cn
}
}


\maketitle

\begin{abstract}
Perceiving the complex driving environment precisely is crucial to the safe operation of autonomous vehicles. With the tremendous advancement of deep learning and communication technology, Vehicle-to-Everything (V2X) cooperative perception has the potential to address limitations in sensing distant objects and occlusion for a single-agent perception system. V2X cooperative perception systems are software systems characterized by diverse sensor types and cooperative agents, varying fusion schemes, and operation under different communication conditions. Therefore, their complex composition gives rise to numerous operational challenges. Furthermore, when cooperative perception systems produce erroneous predictions, the types of errors and their underlying causes remain insufficiently explored.


To bridge this gap, we take an initial step by conducting an empirical study of V2X cooperative perception. To systematically evaluate the impact of cooperative perception on the ego vehicle's perception performance, we identify and analyze six prevalent error patterns in cooperative perception systems. We further conduct a systematic evaluation of the critical components of these systems through our large-scale study and identify the following key findings: 
(1) The LiDAR-based cooperation configuration exhibits the highest perception performance; (2) Vehicle-to-infrastructure (V2I) and vehicle-to-vehicle (V2V) communication exhibit distinct cooperative perception performance under different fusion schemes;
(3) Increased cooperative perception errors may result in a higher frequency of driving violations; (4) Cooperative perception systems are not robust against communication interference when running online.
Our results reveal potential risks and vulnerabilities in critical components of cooperative perception systems. We hope that our findings can better promote the design and repair of cooperative perception systems.



\end{abstract}




\section{Introduction}
\label{sec:introduction}

Autonomous driving systems (ADSs) have attracted significant attention from both industry and academia due to their great potential and long-term value~\cite{DBLP:journals/csur/KhanEMZKAI23}. Among the core components of ADS, the perception module plays a pivotal role, processing and interpreting multi-modal sensor data through deep neural networks (DNNs) to detect and localize obstacles. Recent advancements in deep learning and sensor fusion technologies have substantially improved the accuracy and robustness of perception systems in autonomous vehicles~\cite{DBLP:journals/ijon/WenJ22,DBLP:journals/spm/LiI20}. Despite these remarkable developments, single-agent perception systems often produce inaccurate or incomplete results due to inherent single-view limitations, which can significantly compromise the safety and performance of ADS. Unfortunately, despite extensive testing of ADS on public roads for tens of thousands of hours~\cite{test}, these limitations can still result in incorrect system behavior or even severe accidents~\cite{crash1,crash2}.



Vehicle-to-everything (V2X) cooperative perception technology has emerged as a promising solution to address the inherent limitations of single-agent perception systems, particularly in detecting occluded objects and long-range obstacles~\cite{DBLP:journals/corr/abs-2401-01544,DBLP:journals/sensors/UsinskisMPDB24}. 
By enabling communication among road participants (e.g., connected vehicles and roadside units), V2X facilitates the sharing and fusion of perception information, thereby significantly improving overall perception accuracy. In particular, vehicle-to-vehicle (V2V) and vehicle-to-infrastructure (V2I) systems have received increasing attention.
Major automotive manufacturers and technology leaders, including Ford, Volvo, and Baidu, are heavily investing in V2X platform development, aiming to empower autonomous vehicles with enhanced environmental perception capabilities for safe navigation in complex urban and highway scenarios~\cite{apollo-v2x, tesla-v2x}.

The rapid advancement of V2X cooperative perception systems has introduced a range of challenges and concerns. The primary concern lies in the reliability of the information and the effectiveness of its fusion.
On the one hand, cooperative agents may introduce erroneous or misleading information into the perception pipeline. On the other hand, even when the information provided by cooperative agents is correct, improper fusion with the existing perception pipeline may still result in inaccurate perception outcomes. As a highly complex system, V2X cooperative perception integrates multiple critical components, including heterogeneous sensors ~(e.g., LiDAR and camera), diverse cooperating agents (e.g., V2V and V2I), environment understanding, and reliable wireless communication. While such integration can effectively address the performance limitations of standalone perception systems, it also introduces new vulnerabilities. Failures or inconsistencies in any individual component may undermine the robustness of the entire V2X pipeline, potentially leading to inaccurate perception and unsafe decision-making by the ego vehicle. Therefore, it is imperative to conduct comprehensive testing and analysis of each key component to identify potential sources of fragility and to evaluate their impact on the overall reliability of cooperative perception.

\begin{figure}[htbp]
	\centering
    \includegraphics[width=\linewidth]{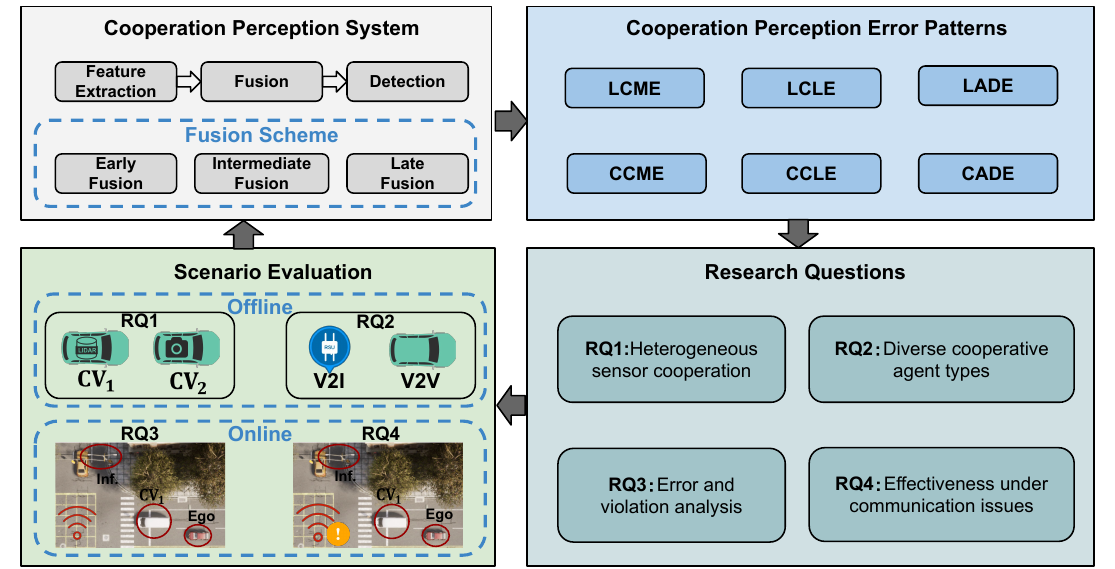}%
	\caption{Workflow of the high-level empirical study on cooperative perception.}
	\label{1-1.pdf}
    \vspace{-15pt}
\end{figure}

To bridge this gap, we take an initial step in this paper by conducting an empirical study of V2X cooperative perception systems. Figure~\ref{1-1.pdf} presents the high-level design and workflow of our study. We begin by performing a comprehensive analysis of V2X cooperative perception systems with different fusion schemes~(detailed in Section~\ref{Heterogeneous V2X Cooperative Perception}) and identifying six distinct cooperative perception error patterns. 
To investigate the performance of key components in the V2X cooperative perception process, we have formulated four research questions investigating the following aspects: (1) the effect of heterogeneous sensor configurations on perception performance (RQ1), (2) the performance differences between V2V and V2I cooperation modes (RQ2), (3) how cooperative perception errors lead to driving violations (RQ3), and (4) the impact of communication issues on system effectiveness (RQ4).
These questions aim to identify critical challenges and highlight potential opportunities for advancing cooperative perception. Finally, we conduct empirical evaluations for each research question and identify several key findings, which we expect could provide feedback to enhance the quality of cooperative perception systems. The detailed findings are summarized as follows: (1) LiDAR-based cooperative configurations achieve the highest perception performance among various sensor setups; (2) V2I and V2V cooperation exhibit distinct performance across fusion schemes, with V2I performs better under the early fusion mechanism and V2V showing superior perception performance under intermediate and late fusion schemes; (3) increased cooperative perception errors could lead to a higher frequency of driving violations; and (4) cooperative perception systems are vulnerable to communication interference. Our findings reveal potential risks and vulnerabilities in critical components of cooperative perception systems, providing insights that could help improve their overall reliability.

The main contributions of this paper are summarized as follows:

\begin{itemize}

\item  To the best of our knowledge, this work presents the first large-scale empirical study of a V2X cooperative perception system. We conduct a comprehensive analysis of error patterns to evaluate the performance of critical components within the cooperative perception systems.

\item  We further discuss existing V2X cooperative perception systems and outline potential future directions, including the advantages of various configurations in heterogeneous environments and opportunities for driving performance enhancement. Moreover, we come up with several novel findings and insights. 
These findings and insights can benefit future research on cooperative perception and support the practical deployment of ADSs.

\item  To support the open science community, we make our source code publicly accessible$\footnote{https://github.com/meng2180/V2X-Cooperative-Perception}$ to facilitate the replication of our study and its application in extensive contexts.

\end{itemize}

This paper represents one of the earliest efforts to investigate cooperative perception systems. This study enables a quantitative investigation of these critical questions and facilitates further research in quality assurance for cooperative perception. In general, V2X cooperative perception plays a key role in enabling ADSs and potentially impacts autonomous driving applications and domains. Given the rapid shift toward a data-driven, intelligent networking era, we believe that early-stage investigation into cooperative perception systems will help practitioners better understand current limitations and develop more robust engineering techniques. Such efforts will pave the way toward designing safer and more reliable V2X-aided ADSs.

\begin{figure}[htbp]\small
	\centering
    \includegraphics[width=0.9\linewidth]{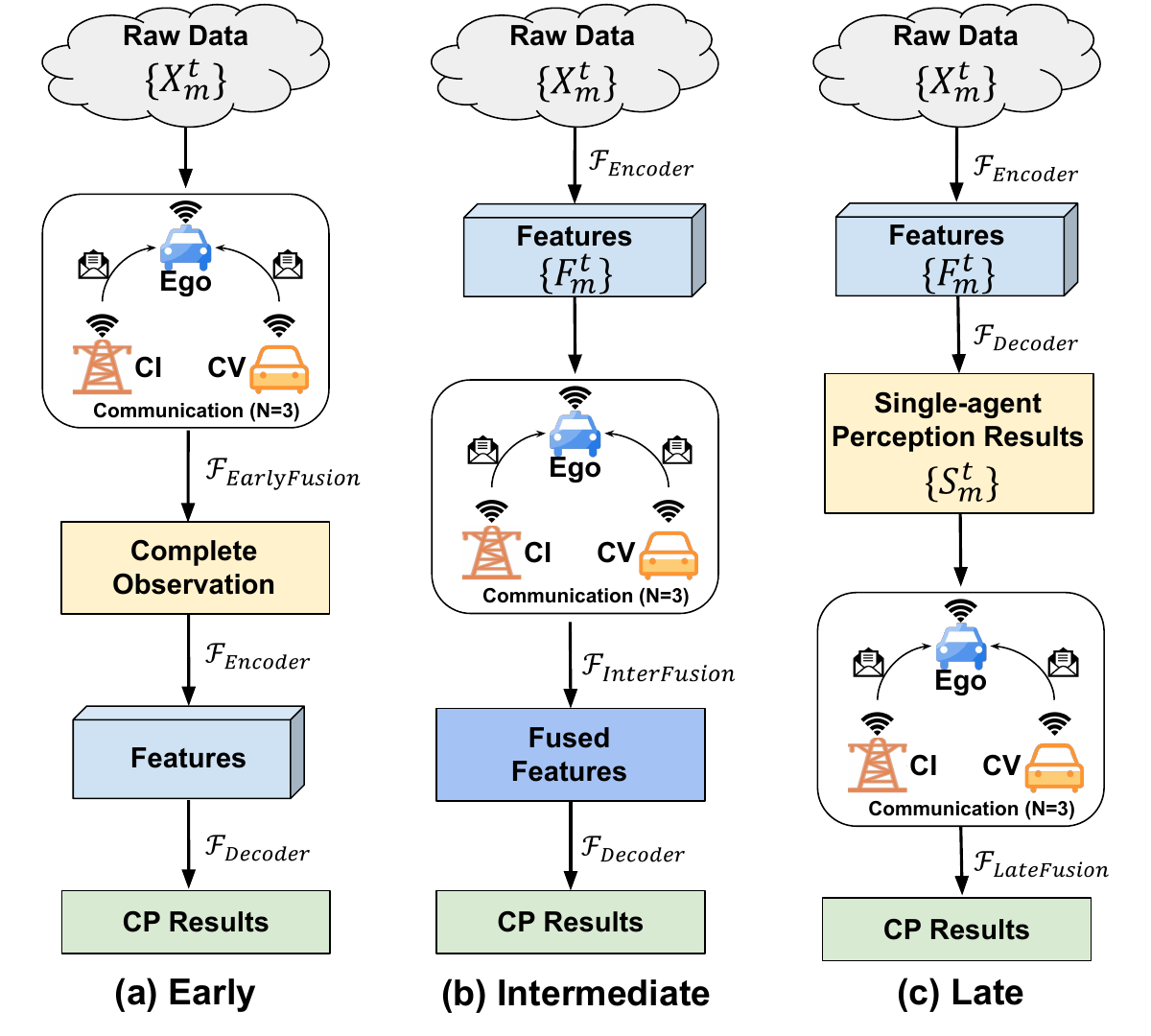}%
        \vspace{-5pt}
	\caption{Overview of cooperative perception schemes for autonomous driving. }
	\label{2-1.pdf}
    \vspace{-25pt}
\end{figure}

\section{Background}

\subsection{Heterogeneous V2X Cooperative Perception}
\label{Heterogeneous V2X Cooperative Perception}

Owing to the inherent limitations of camera and LiDAR sensors, occlusion and long-range perception remain persistent challenges for autonomous systems in single-vehicle settings~\cite{xu2022opv2v}. These limitations may result in critical failures in complex traffic scenarios. Conversely, cooperative perception systems enable multi-vehicle cooperative detection, overcoming the constraints of single-vehicle perception. By utilizing V2X communication, connected vehicles and infrastructure can share sensory data, yielding multiple viewpoints of obstacles and compensating for mutual blind spots~\cite{DBLP:journals/cm/HobertFLAVK15,DBLP:journals/sensors/JungLLS20}.

Multi-agent heterogeneous cooperative perception enables a group of agents to jointly perceive their environment and exchange complementary perception information. 
In real-world scenarios, various operationally and structurally different cooperative agents (e.g., vehicles and roadside units (RSUs)) are typically equipped with different sensors (e.g., LiDAR and cameras), sharing heterogeneous environmental perception information recorded in specific formats (e.g., images and point clouds). The ego vehicle then receives and fuses this heterogeneous information using a specific fusion scheme to accomplish the perception task (e.g., object detection).
The following details the heterogeneous cooperative perception fusion scheme and the cooperative 3D object detection task.

\textbf{Fusion Scheme.} V2X cooperative perception methods can be divided into three categories based on the information fusion stage: \textit{early fusion}, \textit{late fusion}, and \textit{intermediate fusion}~\cite{DBLP:journals/corr/abs-2310-03525}. 
A schematic overview of these cooperative perception schemes is provided in Figure~\ref{2-1.pdf}. For mathematical clarity, consider a scenario with $N$ connected agents including one ego vehicle~(denoted as $E$) and $N-1$ cooperative agents~(denoted as $C_{i}$), where the local raw sensor data observed by the $i$-th agent at time step $t$ is denoted as $X_m^t$ ($m\in \{E, C_{i}\}| i \in 1,2,\ldots,N-1 \}$).
As illustrated in Figure~\ref{2-1.pdf}(a), early fusion involves aggregating raw sensor observations collected by all participating agents, fostering a comprehensive perspective~\cite{DBLP:conf/ccnc/LiYSS23}. The ego receives the raw observation $X_{C_i}^t(i \in\{1, \cdots, N-1\})$. After that, it performs data fusion $\mathcal{F}_{\text {EarlyFusion}}$ on the received raw data with self-raw data and produces a fused complete observation. Then, the ego vehicle uses an encoder $\mathcal{F}_{\text {Encoder }}(\cdot)$ to extract high-level cooperative perception features from its fused complete observation.
Finally, the ego vehicle uses a decoder $\mathcal{F}_{\text {Decoder }}(\cdot)$ to decode the cooperative perception feature and obtain the cooperative perception results. As illustrated in Figure~\ref{2-1.pdf}(b), intermediate fusion shared intermediate features rather than raw data. Firstly, each agent extracts its own feature data $F_m^t$ ($m\in \{E, C_{i}\}| i \in 1,2,\ldots,N-1 \}$) from its raw perception data $X_m^t$. Then, the extracted individual feature data of connected agents is shared with the ego vehicle via V2X communication. After receiving shared feature data $F_{C_i}^t$ from other $N-1$ agents, the ego vehicle uses an intermediate fusion algorithm $\mathcal{F}_{\text {InterFusion }}(\cdot)$ to generate the fused cooperative perception feature data. Finally, the fused cooperative perception feature is fed into the decoder $\mathcal{F}_{\text {Decoder }}(\cdot)$ to obtain the final cooperative perception result. Late fusion involves exchanging individual perceptual outcomes, as depicted in Figure~\ref{2-1.pdf}(c), requiring minimal data transmission. In this case, each agent independently processes its raw perception data $X_m^t$, generates its feature data $F_m^t$, and its final perception result $S_m^t$. After individual processing, the individual perception results from each agent are shared with the ego vehicle. Then, it uses a late fusion algorithm $\mathcal{F}_{\text {LateFusion }}(\cdot)$ to fuse the received individual perception result from all other $N-1$ agents and its own perception result to generate the cooperative perception result.

\textbf{The Preliminaries of Cooperative 3D Object Detection.} 
Cooperative 3D object detection serves as a core perception task for V2X communication systems, allowing ADS to achieve comprehensive environmental awareness. In cooperative 3D object detection, the cooperative perception system takes a multi-view scene data as input, which consists of an ego vehicle’s data $X_{E}^t$ and data $X_{coop}^t=\{X_{C_{i}}^t \}_{i=1}^{n-1}$ from the observation perspectives of $n-1$ cooperative agents. The system then outputs a bounding box for each detected object based on the relative position of the ego vehicle, providing its 3d location $loc = [x, y, z]$, dimensions length $l$, width $w$, height $h$, and orientation through a heading angle $yaw$. 
The accuracy of cooperative 3D object detection is typically evaluated using the Intersection over Union (IoU) metric~\cite{DBLP:conf/iwssip/PadillaNS20}. IoU is computed as $IoU = \text{area}(b_p \cap b_{gt}) / \text{area}(b_p \cup b_{gt})$, where $b_{gt}$ denotes the ground-truth 3D bounding box and $b_p$ represents the predicted 3D bounding box, both projected in bird’s-eye view. This metric quantifies the ratio of the overlapping area to the union area of the predicted and ground-truth boxes. An object is considered successfully detected if the IoU exceeds a predefined threshold $\tau$.

\subsection{Offline and Online Testing}

\begin{figure}[htbp]\small
	\centering
    \includegraphics[width=\linewidth]{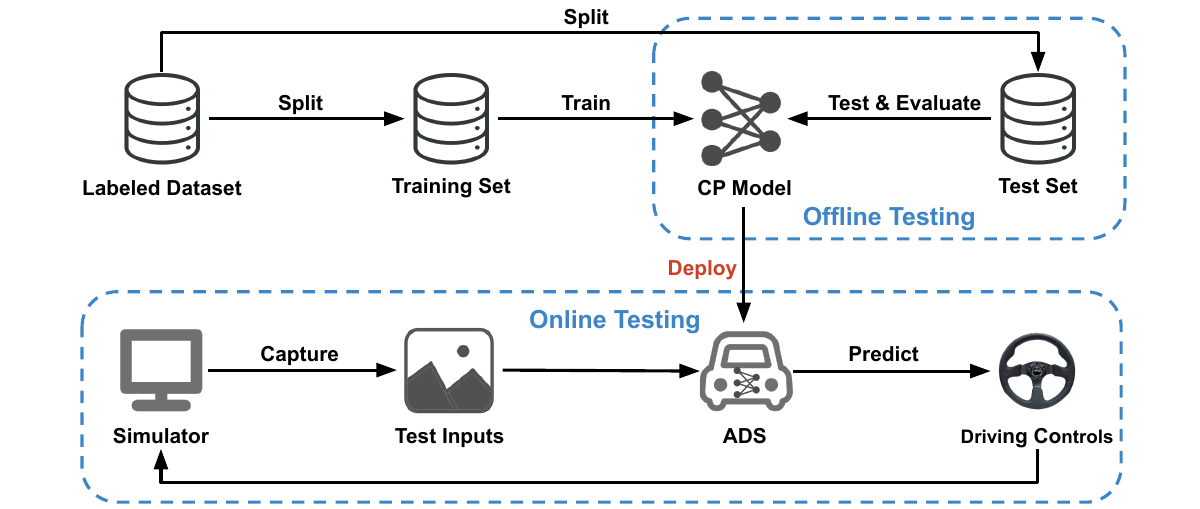}%
        \vspace{-5pt}
	\caption{Offline and online phases of cooperative perception system testing.}
	\label{2-2.pdf}
    \vspace{-15pt}
\end{figure}

Testing has emerged as a well-established and effective approach for assessing the potential risks associated with deploying a software system in real-world scenarios.
Unlike traditional software testing, deep learning-based cooperative perception system testing follows a structured workflow comprising two distinct phases: offline testing and online testing~\cite{DBLP:journals/ese/StoccoPT23,haq2021can}, as illustrated in Figure~\ref{2-2.pdf}. Offline testing constitutes a standard and essential step in the development of cooperative perception systems, typically conducted immediately after system training. Its primary objective is to ensure that the trained system demonstrates sufficient accuracy when applied to a test dataset. 
Following offline testing, online testing is performed by software testers by deploying the cooperative perception system on a specific ADS to examine its interaction with the operational environment under realistic conditions.

Specifically, during offline testing, cooperative perception systems are tested as a unit in open-loop mode. They receive test inputs that can be generated without the participation of the system under test, either manually or automatically. The system's performance is generally assessed by comparing the predicted output bounding boxes with the expected ground truth boxes. Conversely, in online testing, the cooperative perception system is evaluated in a closed-loop environment within a driving simulator. In this phase, test inputs are generated dynamically by the simulated environment, and the ADS's outputs are fed back into the simulator. Online testing evaluates cooperative perception systems by monitoring for safety-critical requirement violations (e.g., collisions or off-road) triggered by the ADS’s output behavior.

\section{Comprehensive Analysis of Cooperative Perception System Bugs}

The purpose of V2X cooperative perception design is to help the ego vehicle have more comprehensive and accurate perception, but it may produce erroneous perception results during operation.
Cooperative errors in complex cooperative perception systems can originate from multiple components, including the ego vehicle, cooperative vehicles or infrastructures, and the information fusion process. This section presents a comprehensive analysis of the error patterns observed in cooperative perception systems and identifies six typical categories. 
Before giving a formal definition of cooperation error patterns, we give a formal representation of V2X-oriented cooperation perception.

\subsection{Formal Representation}

We first formalize the V2X-oriented  cooperative perception system and the ego vehicle perception system.
Consider $n-1$ cooperative agents and an ego vehicle in the scene. Let $X_E^t$ and $X_{C_i}^t$ represent the observation of the ego vehicle and the $i$-th cooperative agent at time step $t$, respectively. The cooperative perception systems $\mathbb{CP}$ can be represented as: 
$$\Phi_{CP}(X_E^t,\{X_{C_i}^t\}_{i=1}^{n-1})=\{(b_i^{CP})|i\in [n_{CP}]\}$$
Similarly, the ego vehicle perception system can be represented as:
$$\Phi_E(X_E^t)=\{(b_i^E)|i\in [n_E]\}$$
,where $x \in \{E, CP\}$ and $\Phi_x$ denotes the perception network of the corresponding perception system, $b_i^x$ represents the $i$-th 3D bounding box in the perception results. 
Let $n_x$ denote the number of 3D bounding boxes detected by the perception system. 
We define the index set of these bounding boxes as $[n_x] = \{1, 2, \ldots, n_x\}$. 
Similarly, let $b_i^T$ denote the $i$-th ground-truth bounding box and $n_T$ the total number of ground-truth 3D bounding boxes. 
We use $B_x = \bigcup_{i=1}^{n_x} b_i^x, \; x \in \{T, E, CP\}$ to represent the set of all $b_i^x$.

\subsection{Cooperative Perception Error Patterns}

The primary purpose of the V2X cooperative perception system is to assist the autonomous driving ego vehicle in achieving more accurate perception. If V2X cooperation fails to improve or even degrades the ego vehicle’s performance, it is considered a cooperative perception error. Based on this principle, we performed pattern mining on an existing cooperative perception dataset, systematically comparing ego perception results with cooperative perception results in identical scenes. This analysis revealed two major error categories: misleading cooperative errors (denoted as LE), where cooperation negatively affects otherwise correct ego perception results, and miscorrected cooperative errors (denoted as CE), where cooperation fails to resolve existing ego perception weaknesses such as occlusions or long-distance detections.

\begin{figure}[htbp]\small
	\centering
    \includegraphics[width=\linewidth]{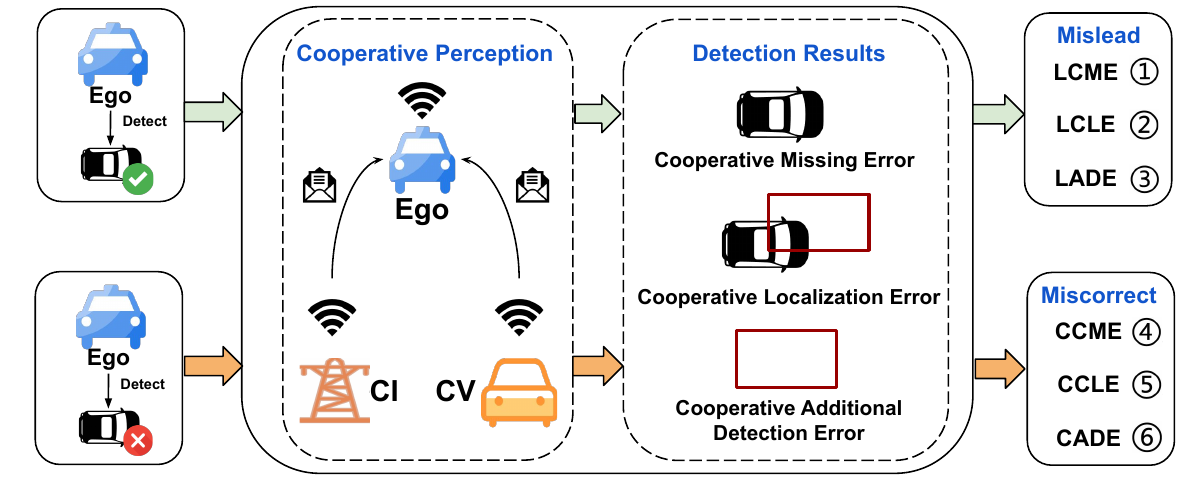}%
        \vspace{-5pt}
	\caption{Graphical representation of error patterns in cooperative perception systems.}
	\label{3-1.pdf}
    \vspace{-15pt}
\end{figure}

In this paper, we focus on the task of cooperative 3D object detection, which serves as a representative task for cooperative perception. Note that, in single-agent 3D detection tasks, error types are typically categorized as missing errors, localization errors, and additional detection errors, as established in prior research~\cite{DBLP:conf/icse/GaoW000X24}. Missing errors occur when the perception system fails to detect an object; localization errors occur when producing an inaccurately localized perception bounding box of the object; and additional detection errors occur when the system predicts objects that do not exist in the scene. Hence, as shown in Figure~\ref{3-1.pdf}, each cooperative perception error category (i.e. misleading cooperative errors (LE) and miscorrected cooperative errors (CE)), can be further subdivided according to these detection error types, resulting in six distinct cooperative perception error subcategories: misleading cooperative missing error (LCME), misleading cooperative localization error (LCLE), misleading cooperative additional detection error (LADE), miscorrected cooperative missing error (CCME), miscorrected cooperative localization error (CCLE), and miscorrected cooperative additional detection error (CADE).
For clarity, the first letter (i.e., L or C) in each error abbreviation indicates the cooperative perception error category (CE or LE), while the remaining letters specify the corresponding 3D object detection error type. In the following, we provide a detailed description of these error patterns.

\subsubsection{Misleading Cooperative Missing Error~(LCME)}
A misleading cooperative missing error occurs (1) when an ego vehicle accurately perceives its objects during operation, and (2) the cooperative perception system may mislead the ego vehicle into losing the perception bounding box of the object due to inaccurate shared information or faulty fusion logic.

\noindent \textbf{Definition.} Consider a cooperative perception system $\mathbb{CP}$.  $\mathbb{CP}$ makes misleading cooperative missing error if
\begin{equation}\small
\begin{aligned}
      \exists b_{i}^T\in B_T, 
s.t.~D_E(b_{i}^T) \wedge \forall b_{j}^{CP}\in B_{CP}, IoU(b_{i}^T, b_{j}^{CP})= 0
\end{aligned}
\nonumber
\end{equation}
, where $D_E$ is the criteria to decide whether the ego vehicle system successfully matches the ground-truth bounding box $b_{i}^T$. $\forall b_{j}^{CP}\in B_{CP}, IoU(b_{i}^T, b_{j}^{CP})= 0$ indicates that, for any predicted detection box $b_{j}^{CP}$ of $\mathbb{CP}$, the IoU between the ground-truth bounding box $b_{i}^T$ of the $i$-th object and $b_{j}^{CP}$ is zero. This condition indicates that the cooperative perception system $\mathbb{CP}$ has entirely missed the detection of the $i$-th object.

\subsubsection{Misleading Cooperative Localization Error~(LCLE)}
A misleading cooperative localization error occurs (1) when an ego vehicle accurately perceives its objects during operation, and (2) the cooperative perception system may mislead the ego vehicle into producing an inaccurately localized perception bounding box of the object due to inaccurate shared information or faulty fusion logic.

\noindent \textbf{Definition.} Consider a cooperative perception system $\mathbb{CP}$.  $\mathbb{CP}$ makes misleading cooperative localization error if
\begin{equation}\small
\begin{aligned}
      \exists b_{i}^T\in B_T, 
s.t.~D_E(b_{i}^T) \wedge \exists b_{j}^{CP}\in B_{CP}, 0<IoU(b_{i}^T, b_{j}^{CP})\leq \tau
\end{aligned}
\nonumber
\end{equation}
, where $\exists b_{j}^{CP}\in B_{CP}, 0<IoU(b_{i}^T, b_{j}^{CP})\leq \tau$ indicates that there exists a predicted detection box $b_{j}^{CP}$ from the cooperative perception system $\mathbb{CP}$ such that the ground-truth bounding box $b_{i}^T$ of the $i$-th object partially overlaps with $b_{j}^{CP}$, but the overlap falls below the threshold $\tau$. This condition indicates that the detection box $b_{j}^{CP}$ predicted by the cooperative perception system $\mathbb{CP}$ has inaccurately localized the $i$-th object.

\subsubsection{Misleading Cooperative Additional Detection Error~(LADE)}
Additional detection refers to the case that the system treats an arbitrary region without objects as an “object”. A misleading cooperative additional detection error occurs (1) when the ego vehicle makes the correct decision about an object that does not exist in the real world during operation, and (2) the cooperative perception system may mislead the ego vehicle into producing an additional detection of a bounding box due to inaccurate shared information or faulty fusion logic. 

\noindent \textbf{Definition.} Consider a cooperative perception system $\mathbb{CP}$.  $\mathbb{CP}$ makes a misleading cooperative additional detection error if
\begin{equation}\small
\begin{aligned}
      \exists b_{j}^{CP}\in B_{CP}, 
s.t. \forall b_{m}^{E}\in B_{E}, IoU(b_{m}^E, b_{j}^{CP})= 0 \\ \wedge \forall b_{i}^{T}\in B_{T}, IoU(b_{i}^T, b_{j}^{CP})= 0
\end{aligned}
\nonumber
\end{equation}
, LADE is considered to occur when the predicted box $b_{j}^{CP}$ generated by $\mathbb{CP}$ neither overlaps with any predicted box $b_{m}^{E}$ of the ego vehicle nor with any ground-truth bounding box $b_{i}^T$ of the object to be detected. The first condition indicates that the ego vehicle does not produce an erroneous prediction similar to $b_{j}^{CP}$, while the second condition indicates that the cooperative perception system $\mathbb{CP}$ makes an erroneous prediction by generating an additional detection box at a location where no obstacle actually exists.

\subsubsection{Miscorrected Cooperative Missing Error~(CCME)}
A miscorrected cooperative missing error occurs (1) when an ego vehicle inaccurately perceives its objects during operation, and (2) the cooperative perception system incorrectly completes or corrects the ego vehicle’s perception results to miss the detection of the object.

\noindent \textbf{Definition.} Consider a cooperative perception system $\mathbb{CP}$.  $\mathbb{CP}$ makes miscorrected cooperative missing error if
\begin{equation}\small
\begin{aligned}
      \exists b_{i}^T\in B_T, 
s.t.~\neg D_E(b_{i}^T) \wedge \forall b_{j}^{CP}\in B_{CP}, IoU(b_{i}^T, b_{j}^{CP})= 0
\end{aligned}
\nonumber
\end{equation}
, where $\neg D_E$ is the criteria to decide whether the ego vehicle system unsuccessfully matches the ground-truth bounding box $b_{i}^T$. The distinction between LCME and CCME lies solely in whether the ego vehicle’s perception system successfully detects the object.

\subsubsection{Miscorrected Cooperative Localization Error~(CCLE)}
A miscorrected cooperation localization error occurs (1) when an ego vehicle inaccurately perceives its objects during operation, and (2) the cooperative perception system incorrectly completes or corrects the ego vehicle’s perception results to produce an inaccurately localized perception bounding box of the object.

\noindent \textbf{Definition.} Consider a cooperative perception system $\mathbb{CP}$.  $\mathbb{CP}$ makes miscorrected cooperative localization error if
\begin{equation}\small
\begin{aligned}
      \exists b_{i}^T\in B_T, 
s.t.~\neg D_E(b_{i}^T) \wedge \exists b_{j}^{CP}\in B_{CP}, 0<IoU(b_{i}^T, b_{j}^{CP})\leq \tau
\end{aligned}
\nonumber
\end{equation}
, where $\neg D_E$ is the criteria to decide whether the ego vehicle system unsuccessfully matches the ground-truth bounding box $b_{i}^T$. The difference between LCLE and CCLE lies only in whether the ego vehicle's perception system successfully detects the object.

\subsubsection{Miscorrected Cooperative Additional Detection Error~(CADE)}
A miscorrected cooperative additional detection error occurs (1) when the ego vehicle produces an additional detection of a bounding box during operation, and (2) the cooperative perception system fails to correct this erroneous detection, instead producing a false detection bounding box near the original additional detection location.

\noindent \textbf{Definition.} Consider a cooperative perception system $\mathbb{CP}$.  $\mathbb{CP}$ makes a miscorrected cooperative additional detection error if
\begin{equation}\small
\begin{aligned}
      \exists b_{j}^{CP}\in B_{CP}, 
s.t. \exists b_{m}^{E}\in B_{E}, IoU(b_{m}^E, b_{j}^{CP})> 0 \\ \wedge \forall b_{i}^{T}\in B_{T}, IoU(b_{i}^T, b_{j}^{CP})= 0
\end{aligned}
\nonumber
\end{equation}
Unlike LADE, CADE occurs when the prediction box $b_{m}^{E}$ of the ego vehicle overlaps with the prediction box $b_{j}^{CP}$ generated by $\mathbb{CP}$. This indicates that the cooperative perception system $\mathbb{CP}$ generates additional detection boxes without correcting the erroneous prediction of the ego vehicle.

\section{Empirical Study Design}

\subsection{Research Questions}

\textbf{RQ1. How does equipping cooperative agents with heterogeneous sensors affect the performance of cooperative perception systems?} 
Heterogeneous environments that integrate different sensor types may introduce new challenges for cooperative perception.
RQ1 aims to evaluate the performance of agents equipped with diverse sensor configurations when cooperating with the ego vehicle. In this RQ, we focus on three cooperation scenarios: LiDAR-based cooperation, camera-based cooperation, and hybrid multimodal (LiDAR and camera) sensor cooperation. Specifically, in the first two cooperative scenarios, all agents are equipped with the same sensor type. In the final multi-modal scenario, participants comprise a heterogeneous mix of LiDAR-equipped and camera-equipped agents, with the ego vehicle randomly assigned one of the sensor configurations in each trial.

\textbf{RQ2. What are the differences in cooperative perception performance between V2V and V2I cooperation modes?}
The cooperative perception system comprises multiple types of intelligent agents, such as vehicles and infrastructure, that contribute to environmental perception. Investigating the contribution of different types of collaborative agents to the ego vehicle’s perception performance is of significant importance. In this research question, we focus on quantifying the differences in V2I and V2V cooperative perception performance under different fusion schemes.

\textbf{RQ3. What is the relationship between cooperative perception errors and driving violations?}
To assess the potential impact of cooperative perception errors on driving violations, we first integrate the cooperative perception system into the ADS and run the integrated ADS in the simulator. This setup allows us to evaluate the system’s environmental perception performance and to identify driving violations, such as collisions, during online testing. 
To investigate the correlation between cooperative perception system errors and ADS violations, we compare the number of cooperative perception errors observed in violation scenarios and non-violation scenarios. For ADS violation scenarios, we further investigate the distribution of cooperative perception errors over the entire duration from ADS startup to the point of violation.

\textbf{RQ4. Do communication issues during online operations diminish the effectiveness of the cooperative perception system?}
During online operation, cooperative perception systems are inevitably affected by real-world communication interferences. Communication latency~(CL) and pose error~(PE) are two typical types of such interference. In practical scenarios, limited communication bandwidth~\cite{DBLP:conf/eccv/WangMLYZU20} or transmission failures~\cite{DBLP:conf/eccv/XuXTXYM22} may introduce delays in data exchange between participants, resulting in communication latency. Similarly, obstacles or GPS signal interference~\cite{liu2021automated} in real-world environments can degrade the accuracy of participants' positioning information, leading to pose errors. In this study, we investigate whether the performance of ADS equipped with cooperative perception capabilities is impacted by these two common forms of communication interference.

\subsection{Datasets and Platform}
\label{Datasets and Platform}

To address RQ1, we employ the widely adopted OPV2V dataset~\cite{xu2022opv2v}, which has been extensively used in heterogeneous cooperative perception research. OPV2V is a large-scale, publicly available simulated dataset designed for V2V perception. It comprises over 70 diverse scenarios, 11,464 frames, and 232,913 annotated 3D vehicle bounding boxes, collected from eight towns in the CARLA simulator~\cite{DBLP:conf/corl/DosovitskiyRCLK17} as well as a digital reconstruction of Culver City, Los Angeles. For RQ2, we require a dataset that encompasses both cooperative vehicles and cooperative infrastructure within cooperative driving scenarios. We employ the V2XSet dataset~\cite{DBLP:conf/eccv/XuXTXYM22} collected from CARLA, an open dataset for V2X cooperative perception that incorporates both cooperative vehicles and infrastructure across its training and testing scenarios. 
For RQ3 and RQ4, we use the CARLA-based testing platform V2Xverse~\cite{liu2025towards}, a comprehensive simulation environment for online collaborative autonomous driving. 
V2Xverse provides an end-to-end pipeline that includes a multi-agent driving dataset generation scheme and a codebase for deploying a full-stack cooperative driving system. Specifically, it supports constructing cooperative perception scenarios by adding connected vehicles and roadside units (RSUs), equipping them with sensors (e.g., cameras or LiDAR), and sharing their perception information with the ego vehicle. We evaluate the ADS equipped with cooperative perception on the default 32 CARLA Town05 routes, retaining only those routes where the ADS shows no violations under perfect perception to ensure that any observed failures are attributable to the cooperative perception system.

\subsection{Cooperative Perception Systems}
To evaluate system performance in heterogeneous environments, we conduct offline experiments (RQ1 \& RQ2) on six cooperative perception systems employing different fusion schemes. These include one early fusion system, one late fusion system, and four state-of-the-art intermediate fusion systems~(DiscoNet~\cite{DBLP:conf/nips/LiRWCFZ21}, V2X-ViT~\cite{DBLP:conf/eccv/XuXTXYM22}, AttFusion~\cite{xu2022opv2v}, F-Cooper~\cite{DBLP:conf/edge/ChenMTGYF19}).
All cooperative perception systems implemented LiDAR, camera, and multi-sensor~(LiDAR and camera) cooperative driving sensor configurations based on HEAL~\cite{DBLP:conf/iclr/LuHZWWC24}. For online performance evaluation (RQ3 \& RQ4), we deploy both early and late fusion systems alongside two representative intermediate fusion architectures (V2X-ViT and F-Cooper) selected from our offline testing systems.

\subsection{Implementation Details}

In all experiments, we set the IoU threshold $\tau$ to 0.5, consistent with prior research~\cite{DBLP:conf/issta/GuoG00LGSF24}. For the online testing experiments, the cooperative agent used V2Xverse’s default test configuration, which enables cooperative perception with RSUs. Specifically, RSUs are strategically positioned along the roadside in proximity to the ego vehicle. To ensure that the ego vehicle consistently has at least one RSU available for valid collaboration, a new RSU is deployed every five seconds as the ego vehicle moves, positioned on the right side of the road, 12 meters ahead of the ego vehicle. For RQ4, the transmission delay in V2X communication varies from 500 ms to 1000 ms. To simulate pose error, we introduce translational offsets along specific axes. 
To simulate pose errors, we introduce translational offsets along the x- and y-axes and rotational offsets around the z-axis to the cooperative agent. The translation offsets for both the x-axis and y-axis are set within the range of 0 to 0.6 meters, while the rotation angle around the z-axis is varied between 0 and 0.6 degrees. We randomly select parameters within the range for two abnormal communication conditions. For details on the experimental environment, please refer to the supplementary website~\cite{website}.

\subsection{Evaluation Metrics and Baseline}

\textbf{Evaluation Metrics.} To evaluate the perception performance of the system in cooperative 3D object detection, we first apply the Intersection over Union (IoU) metric (Section~\ref{Heterogeneous V2X Cooperative Perception}) to identify successful detections. We then assess detection performance using Average Precision (AP), a standard evaluation metric in V2X cooperative perception research~\cite{DBLP:conf/cvpr/Xu0LLZTMXDSYZM23,DBLP:conf/issta/GuoG00LGSF24}, which is defined as follows:
\begin{equation}\small
\begin{aligned}
\left.\mathrm{AP}\right|_{R}=\frac{1}{|R|} \sum_{r \in R} \rho_{\text {interp }}(r)
\end{aligned}
\nonumber
\end{equation}
For fair comparisons, we adopt the 11 recall positions from the Pascal VOC benchmark~\cite{everingham2010pascal}, denoted as $R_{11}=\{0, 0.1, 0.2,\ldots, 1\}$. The interpolation function, $\rho_{\text{interp}}(r)=\max _{r^{\prime}: r^{\prime} \geq r} \rho\left(r^{\prime}\right)$, is employed, where $\rho(r)$ represents the precision at recall $r$. A higher AP reflects better performance of cooperative perception systems.

To evaluate the driving performance of the ADS equipped with cooperative perception systems, we use the driving score defined in the CARLA Autonomous Driving Leaderboard 1.0~\cite{leaderboard} as the evaluation metric. Specifically, the driving score (DS) is computed as $DS = R_k \times P_k$, where $R_k$ is the percentage of the $k$-th route completed before the first violation of ADS, and $P_k$ is the corresponding infraction penalty.  
The penalty rate $P_k$ is defined as:
\begin{equation}
\begin{aligned}
P_k=\prod_{j_\in J} p_j^{n_j}
\end{aligned}
\nonumber
\end{equation}
where $p_j$ is the penalty rate for an incident type $j$ from a given set of incident types $J$, and $n_j$ is the number of occurrences of this type. Incident types refer to infractions, such as collisions with other vehicles, as defined in the CARLA Autonomous Driving Leaderboard~\cite{leaderboard}.

To further evaluate the driving performance of the ADS equipped with cooperative perception systems in greater detail, we employ two additional metrics: collision rate (CR) and route completion rate (RCR). The CR refers to the ratio of the number of routes with collisions to the total number of test routes, while the RCR refers to the average percentage of route completion across all test routes.

\begin{table}[htbp]
  \centering
  \vspace{-5pt}
  \caption{Cooperative error pattern analysis of cooperative perception systems under heterogeneous sensor configurations.}
  \vspace{-5pt}
    \setlength{\tabcolsep}{1.1mm}
    \begin{tabular}{lllllllll}
    \hline
    \multicolumn{1}{c|}{\textbf{Systems}} & \multicolumn{1}{c|}{\textbf{Sensor}} & \multicolumn{1}{c|}{\textbf{LCME}} & \multicolumn{1}{c|}{\textbf{LCLE}} & \multicolumn{1}{c|}{\textbf{LADE}} & \multicolumn{1}{c|}{\textbf{CCME}} & \multicolumn{1}{c|}{\textbf{CCLE}} & \multicolumn{1}{c|}{\textbf{CADE}} & \multicolumn{1}{c}{\textbf{AP}} \bigstrut\\
    \hline
    \multicolumn{1}{c|}{\textbf{Early Fusion}} & \multicolumn{1}{c|}{\textbf{L}} & \multicolumn{1}{c|}{\textbf{0.0 }} & \multicolumn{1}{c|}{\textbf{0.0 }} & \multicolumn{1}{c|}{\textbf{0.0 }} & \multicolumn{1}{c|}{\textbf{0.3 }} & \multicolumn{1}{c|}{\textbf{0.0 }} & \multicolumn{1}{c|}{\textbf{1.2 }} & \multicolumn{1}{c}{\textbf{0.96}} \bigstrut\\
    \hline
    \multicolumn{1}{c|}{\multirow{3}[6]{*}{\textbf{Late Fusion}}} & \multicolumn{1}{c|}{\textbf{L}} & \multicolumn{1}{c|}{\textbf{0.0 }} & \multicolumn{1}{c|}{\textbf{0.0 }} & \multicolumn{1}{c|}{\textbf{1.1 }} & \multicolumn{1}{c|}{\textbf{0.3 }} & \multicolumn{1}{c|}{\textbf{0.1 }} & \multicolumn{1}{c|}{\textbf{1.2 }} & \multicolumn{1}{c}{\textbf{0.96}} \bigstrut\\
\cline{2-9}    \multicolumn{1}{c|}{} & \multicolumn{1}{c|}{\textbf{C}} & \multicolumn{1}{c|}{0.0 } & \multicolumn{1}{c|}{2.0 } & \multicolumn{1}{c|}{2.5 } & \multicolumn{1}{c|}{0.6 } & \multicolumn{1}{c|}{4.0 } & \multicolumn{1}{c|}{2.0 } & \multicolumn{1}{c}{0.73} \bigstrut\\
\cline{2-9}    \multicolumn{1}{c|}{} & \multicolumn{1}{c|}{\textbf{M}} & \multicolumn{1}{c|}{0.0 } & \multicolumn{1}{c|}{1.2 } & \multicolumn{1}{c|}{1.6 } & \multicolumn{1}{c|}{0.8 } & \multicolumn{1}{c|}{1.6 } & \multicolumn{1}{c|}{1.3 } & \multicolumn{1}{c}{0.87} \bigstrut\\
    \hline
    \multicolumn{1}{c|}{\multirow{3}[6]{*}{\textbf{F-Cooper}}} & \multicolumn{1}{c|}{\textbf{L}} & \multicolumn{1}{c|}{\textbf{0.0 }} & \multicolumn{1}{c|}{\textbf{1.0 }} & \multicolumn{1}{c|}{\textbf{0.3 }} & \multicolumn{1}{c|}{\textbf{0.3 }} & \multicolumn{1}{c|}{\textbf{0.4 }} & \multicolumn{1}{c|}{\textbf{0.9 }} & \multicolumn{1}{c}{\textbf{0.87}} \bigstrut\\
\cline{2-9}    \multicolumn{1}{c|}{} & \multicolumn{1}{c|}{\textbf{C}} & \multicolumn{1}{c|}{0.1 } & \multicolumn{1}{c|}{1.1 } & \multicolumn{1}{c|}{0.6 } & \multicolumn{1}{c|}{2.0 } & \multicolumn{1}{c|}{4.1 } & \multicolumn{1}{c|}{0.9 } & \multicolumn{1}{c}{0.49} \bigstrut\\
\cline{2-9}    \multicolumn{1}{c|}{} & \multicolumn{1}{c|}{\textbf{M}} & \multicolumn{1}{c|}{0.0 } & \multicolumn{1}{c|}{1.1 } & \multicolumn{1}{c|}{0.8 } & \multicolumn{1}{c|}{0.8 } & \multicolumn{1}{c|}{2.5 } & \multicolumn{1}{c|}{1.1 } & \multicolumn{1}{c}{0.68} \bigstrut\\
    \hline
    \multicolumn{1}{c|}{\multirow{3}[6]{*}{\textbf{DiscoNet}}} & \multicolumn{1}{c|}{\textbf{L}} & \multicolumn{1}{c|}{\textbf{0.0 }} & \multicolumn{1}{c|}{\textbf{0.1 }} & \multicolumn{1}{c|}{\textbf{0.1 }} & \multicolumn{1}{c|}{\textbf{0.5 }} & \multicolumn{1}{c|}{\textbf{0.3 }} & \multicolumn{1}{c|}{\textbf{0.6 }} & \multicolumn{1}{c}{\textbf{0.92}} \bigstrut\\
\cline{2-9}    \multicolumn{1}{c|}{} & \multicolumn{1}{c|}{\textbf{C}} & \multicolumn{1}{c|}{0.1 } & \multicolumn{1}{c|}{0.8 } & \multicolumn{1}{c|}{0.2 } & \multicolumn{1}{c|}{2.5 } & \multicolumn{1}{c|}{2.8 } & \multicolumn{1}{c|}{1.2 } & \multicolumn{1}{c}{0.51} \bigstrut\\
\cline{2-9}    \multicolumn{1}{c|}{} & \multicolumn{1}{c|}{\textbf{M}} & \multicolumn{1}{c|}{0.0 } & \multicolumn{1}{c|}{0.5 } & \multicolumn{1}{c|}{0.7 } & \multicolumn{1}{c|}{0.9 } & \multicolumn{1}{c|}{2.7 } & \multicolumn{1}{c|}{1.5 } & \multicolumn{1}{c}{0.7} \bigstrut\\
    \hline
    \multicolumn{1}{c|}{\multirow{3}[6]{*}{\textbf{AttFusion}}} & \multicolumn{1}{c|}{\textbf{L}} & \multicolumn{1}{c|}{\textbf{0.0 }} & \multicolumn{1}{c|}{\textbf{0.1 }} & \multicolumn{1}{c|}{\textbf{0.1 }} & \multicolumn{1}{c|}{\textbf{0.4 }} & \multicolumn{1}{c|}{\textbf{0.3 }} & \multicolumn{1}{c|}{1.3 } & \multicolumn{1}{c}{\textbf{0.93}} \bigstrut\\
\cline{2-9}    \multicolumn{1}{c|}{} & \multicolumn{1}{c|}{\textbf{C}} & \multicolumn{1}{c|}{0.1 } & \multicolumn{1}{c|}{0.8 } & \multicolumn{1}{c|}{0.1 } & \multicolumn{1}{c|}{2.4 } & \multicolumn{1}{c|}{3.1 } & \multicolumn{1}{c|}{\textbf{0.9} } & \multicolumn{1}{c}{0.53} \bigstrut\\
\cline{2-9}    \multicolumn{1}{c|}{} & \multicolumn{1}{c|}{\textbf{M}} & \multicolumn{1}{c|}{0.0 } & \multicolumn{1}{c|}{0.4 } & \multicolumn{1}{c|}{0.7 } & \multicolumn{1}{c|}{1.8 } & \multicolumn{1}{c|}{3.1 } & \multicolumn{1}{c|}{1.8 } & \multicolumn{1}{c}{0.71} \bigstrut\\
    \hline
    \multicolumn{1}{c|}{\multirow{3}[6]{*}{\textbf{V2XViT}}} & \multicolumn{1}{c|}{\textbf{L}} & \multicolumn{1}{c|}{\textbf{0.0 }} & \multicolumn{1}{c|}{\textbf{0.0 }} & \multicolumn{1}{c|}{\textbf{0.2 }} & \multicolumn{1}{c|}{\textbf{0.3 }} & \multicolumn{1}{c|}{\textbf{0.1 }} & \multicolumn{1}{c|}{\textbf{0.5 }} & \multicolumn{1}{c}{\textbf{0.96}} \bigstrut\\
\cline{2-9}    \multicolumn{1}{c|}{} & \multicolumn{1}{c|}{\textbf{C}} & \multicolumn{1}{c|}{0.0 } & \multicolumn{1}{c|}{0.7 } & \multicolumn{1}{c|}{0.6 } & \multicolumn{1}{c|}{2.0 } & \multicolumn{1}{c|}{3.1 } & \multicolumn{1}{c|}{0.9 } & \multicolumn{1}{c}{0.6} \bigstrut\\
\cline{2-9}    \multicolumn{1}{c|}{} & \multicolumn{1}{c|}{\textbf{M}} & \multicolumn{1}{c|}{0.0 } & \multicolumn{1}{c|}{0.4 } & \multicolumn{1}{c|}{0.3 } & \multicolumn{1}{c|}{1.7 } & \multicolumn{1}{c|}{2.3 } & \multicolumn{1}{c|}{1.7 } & \multicolumn{1}{c}{0.89} \bigstrut\\
    \hline
    \multicolumn{9}{l}{\textbf{*Current lacks the early fusion design that leverages C and M configurations}} \bigstrut[t]\\
    \multicolumn{9}{l}{\textbf{due to the absence of methods that directly accumulate images.}} \\
    \end{tabular}%
\vspace{-10pt}
  \label{rq1-1}%
\end{table}%

\begin{figure}[htbp]\small
	\centering
    \includegraphics[width=\linewidth]{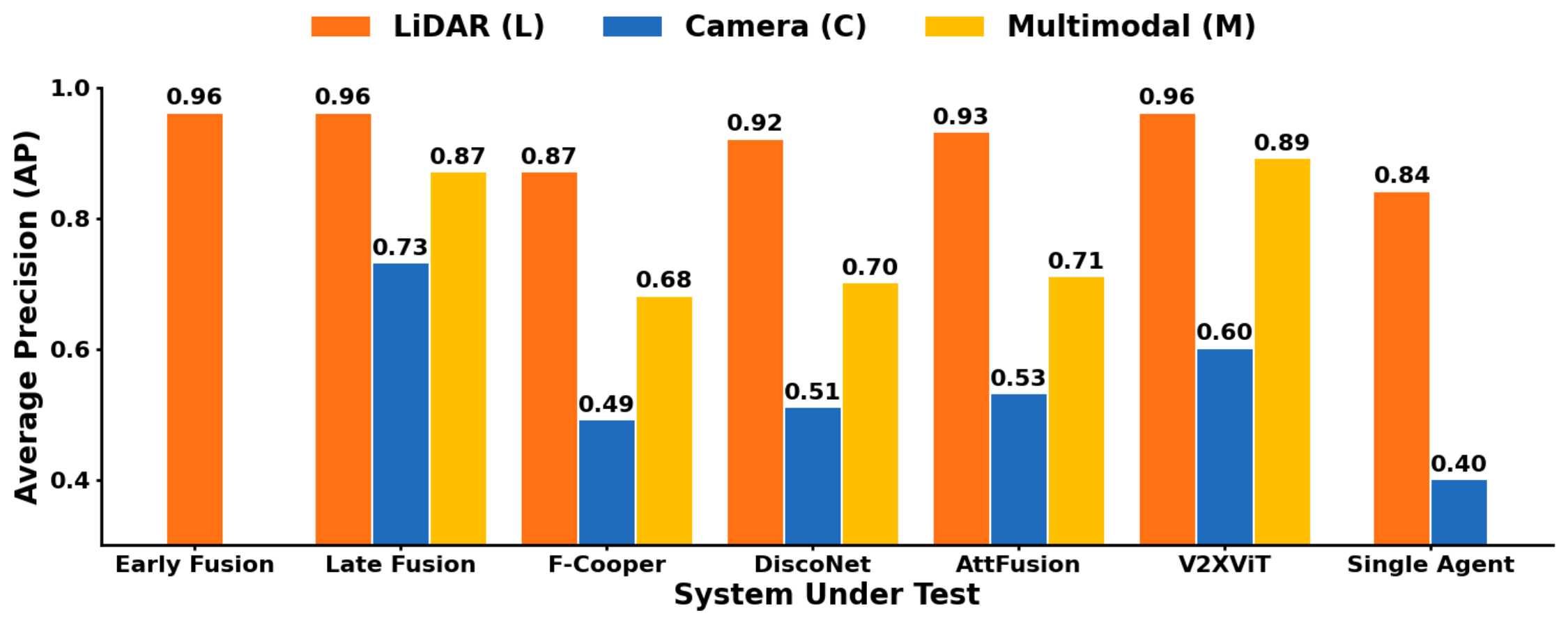}%
        \vspace{-5pt}
	\caption{Average precision of the system under test under heterogeneous sensor configurations.}
	\label{RQ1-single}
    \vspace{-25pt}
\end{figure}

\textbf{Baseline.} For each RQ, we introduce a baseline using a single-agent perception system. This baseline represents the ego vehicle performing environmental understanding independently, without incorporating information from cooperative agents. Specifically, in RQ1, the single-agent system employs either LiDAR or camera sensors for perception; in RQ2, we compare its perception performance with that of cooperative systems using V2V and V2I communication modes. For the online evaluations in RQ3 and RQ4, we compare the performance of the cooperative perception system under both normal and abnormal communication conditions against that of a single-agent perception system.

\section{Results and Findings}

\subsection{RQ1.The LiDAR-based cooperation configuration exhibits the highest perception performance.}

\subsubsection{Experimental Setup}

To evaluate the performance of the cooperative perception system in a heterogeneous sensor environment, we conduct experiments using pre-trained systems under three cooperative configurations: LiDAR-based cooperation (L), camera-based cooperation (C), and multimodal sensor cooperation (M). Each pre-trained system is tested using the original test scenes and the default configuration of the HEAL heterogeneous framework.
We assess the prediction performance of the three sensor configurations using the same set of test data, collect their detection outputs, and compute both the average number of cooperative perception errors per test frame and the overall AP.

\subsubsection{Experimental Results}

Table~\ref{rq1-1} presents an analysis of cooperative error patterns exhibited by cooperative perception systems under heterogeneous sensor configurations, detailing the average number of each error pattern per test frame. The perception performance achieved through LiDAR-based cooperation significantly exceeds that of the other sensor configurations. Specifically, for each tested system configuration, it outperforms multimodal sensor cooperation by an average of 21.7\% and camera-based cooperation by an average of 87.0\%. Moreover, the total number of CCME and CCLE errors in camera-based cooperative perception is substantially higher than in LiDAR-based cooperative perception, leading to a pronounced decline in the AP of the camera-based configuration.
Additionally, we observe that the number of cooperative localization errors is, on average, 122.0\% and 38.8\% greater than the number of cooperative missing errors and additional detection errors, respectively, across all configurations. Furthermore, in most configurations, the number of miscorrected cooperative errors exceeds that of misleading cooperative errors. This indicates that the majority of errors arise from the cooperative perception system's failure to correct inaccurate or incomplete information initially predicted by the ego vehicle.
Figure~\ref{RQ1-single} presents AP of the system under test under heterogeneous sensor configurations. When using the same type of sensors (LiDAR or camera), cooperative perception systems perform better than the single-agent system. Multimodal sensor cooperation outperforms camera-based single-agent perception, while its performance relative to LiDAR-based single-agent perception depends on the specific model architecture. To realize optimal cooperative perception, it is essential to appropriately configure sensor setups and select model architectures suited to heterogeneous environments.


\vspace{-5pt}
\begin{center}
\begin{tcolorbox}[colback=gray!15,
                  colframe=black,
                  width=9cm,
                  arc=1mm, auto outer arc,
                  boxrule=0.5pt,size=title,opacityfill=0.1
                 ]
\textbf{Answer to RQ1:} Compared to camera-based cooperation and multimodal sensor cooperation, using LiDAR for all cooperative agents yields superior performance. Additionally, we find that CCME and CCLE could be the primary factors contributing to the reduction in AP of the other two sensor configurations (C and M).
\end{tcolorbox}
\end{center} 
\vspace{-10pt}

\subsection{RQ2. V2I and V2V communication exhibit distinct cooperative perception performance under different fusion schemes.} 
\subsubsection{Experimental Setup}
To investigate differences in cooperative perception performance between V2V and V2I cooperation
modes, we select cooperative vehicles and infrastructure to collaborate with the ego vehicle. To ensure a fair evaluation of the roles of cooperative vehicles and infrastructure, we randomly select a cooperative agent for each cooperation mode in each scenario.
In addition, to eliminate the influence of sensor heterogeneity, all cooperative agents are equipped with the same type of sensor during the experiment.
Specifically, we conduct experiments using configurations for LiDAR-based cooperation under the original test set. 
We further statistically analyze the effect of different types of cooperative agents on the cooperation of the ego vehicle at different distances. Specifically, we divide the detection targets into three ranges~(refer to~\cite{DBLP:conf/cvpr/Xu0LLZTMXDSYZM23,liu2023towards}) according to the distance relative to the ego vehicle, including 0-30m~(Short), 30-50m~(Middle), and 50-100m~(Long). Then we calculate the cooperative error numbers at each level and AP values.

\begin{figure}[htbp]\small
	\centering
    \includegraphics[width=\linewidth]{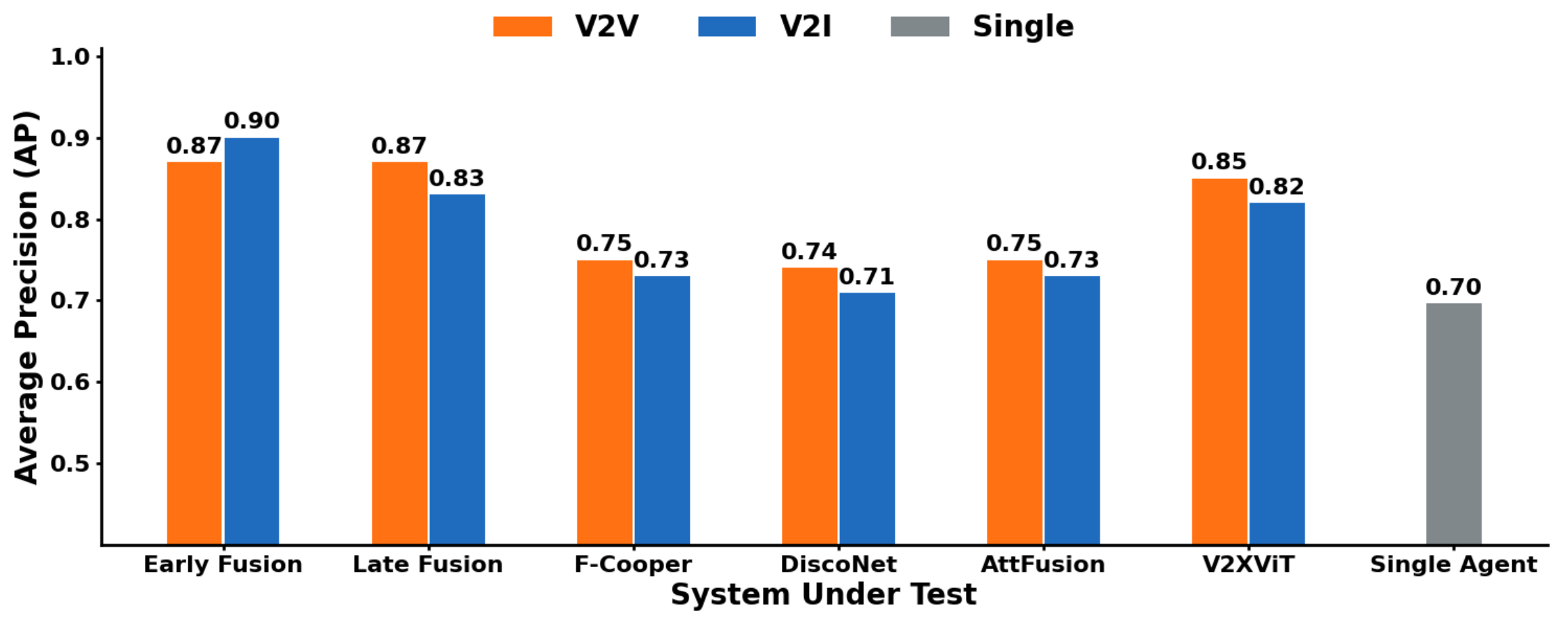}%
        \vspace{-5pt}
	\caption{Average precision of the system under test under different cooperation modes.}
	\label{RQ2-single}
    \vspace{-30pt}
\end{figure}

\subsubsection{Experimental Results}
\begin{figure}[htbp]\small
	\centering
    \includegraphics[width=\linewidth]{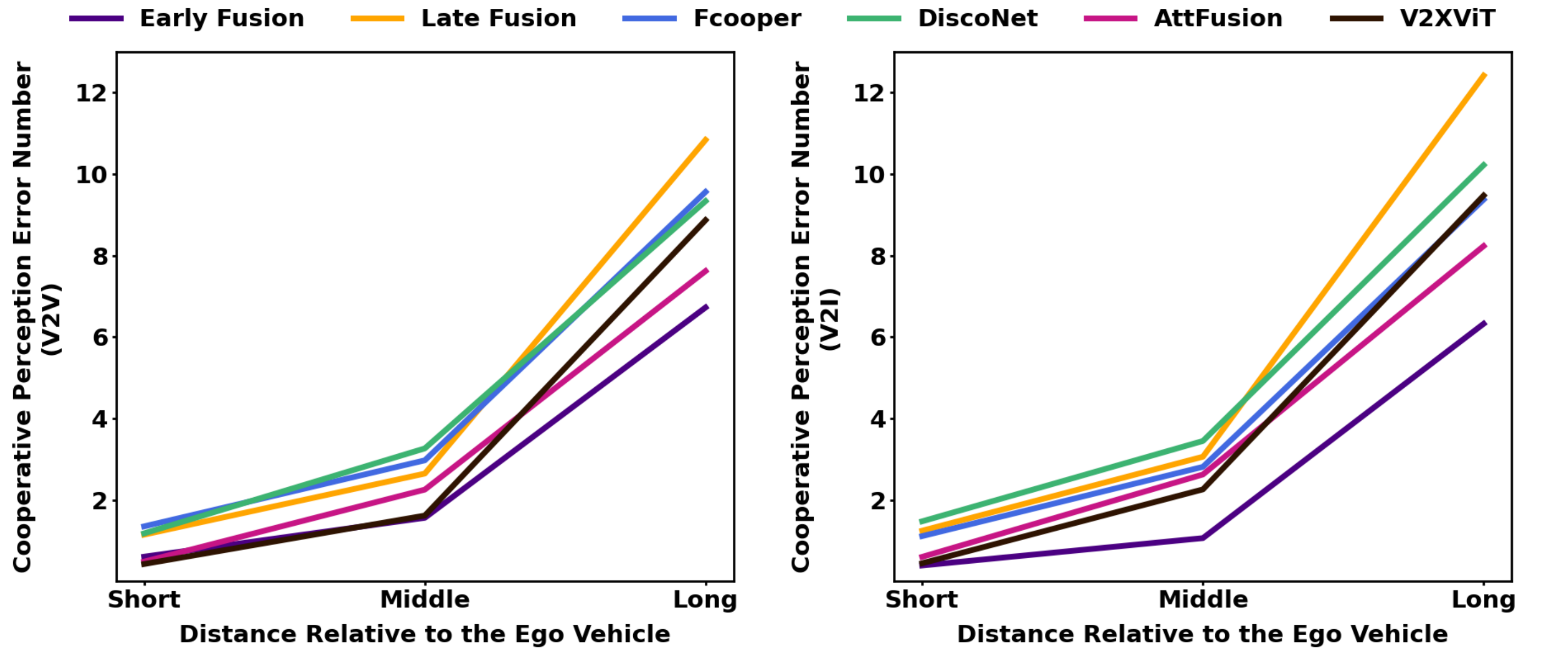}%
        \vspace{-5pt}
	\caption{The cooperative perception error number of the system at varying distances under V2V and V2I cooperation.}
	\label{RQ2-1}
    \vspace{-25pt}
\end{figure}

Figure~\ref{RQ2-single} shows the AP under different cooperation modes. It demonstrates that V2V outperforms V2I under intermediate and late fusion, whereas V2I outperforms V2V under early fusion. Both V2V and V2I surpass single-agent perception, confirming their effectiveness in enhancing perception performance. Figure~\ref{RQ2-1} illustrates the total cooperative perception error number of the system at varying distances under various cooperation modes. We find that the number of cooperative perception errors increases with distance, indicating that the greater the distance between the ego vehicle and the target obstacle, the lower the cooperative perception performance of both the cooperative vehicle and the cooperative infrastructure. For the complete analysis results of six cooperative perception error patterns, please refer to the supplementary website~\cite{website}.

\vspace{-5pt}
\begin{center}
\begin{tcolorbox}[colback=gray!15,
                  colframe=black,
                  width=9cm,
                  arc=1mm, auto outer arc,
                  boxrule=0.5pt,size=title,opacityfill=0.1
                 ]
\textbf{Answer to RQ2:} 
Given the same number of cooperative agents, cooperative vehicles demonstrate superior perception performance under intermediate and late fusion schemes, whereas cooperative infrastructure performs better under the early fusion mechanism.

\end{tcolorbox}
\end{center} 
\vspace{-10pt}

\subsection{RQ3. Increased cooperative perception errors may result in a higher frequency of driving violations.}

\subsubsection{Experimental Setup}
To evaluate the relationship between cooperative perception errors and driving violations during online operation, we connect the ADS equipped with the V2X cooperative perception to a simulator for testing. Each cooperative perception system is evaluated on the V2XVerse online testing platform using the selected routes described above in Section~\ref{Datasets and Platform}. Because autonomous driving accidents caused by perception systems are often precipitated by a sequence of erroneous perceptions within a short time window preceding a collision, and following the similar methodology of prior study~\cite{DBLP:conf/issta/ZhongHGZZR22}, we focus our analysis on the final scene segment associated with each accident. During each route test, we record the following: (1) the number of six predefined errors of the cooperative perception system within the last 10 seconds of the scenario, and (2) whether a violation occurred (i.e., collision with a non-player character (NPC), off-road, or route incomplete). Then we calculate the driving score (DS), collision rate (CR), and average route completion rate (RCR) for all routes.

To further investigate the distribution of perception errors over the entire duration from the initiation of the ADS to the occurrence of a violation, we divide this period into consecutive 5-second intervals. Specifically, we record the number of errors occurring in the 0–5 seconds preceding the violation (denoted as $t_3$), the number of errors in the 5–10 seconds prior to the violation (denoted as $t_2$), and the average number of errors in all earlier intervals beyond 10 seconds before the violation (denoted as $t_1$).

Finally, to quantify the independent effects of different types of cooperative perception errors on driving violations, we employ a multivariable logistic regression model~\cite{DBLP:journals/infsof/CerpaBKV10}, which is commonly used in software engineering to evaluate the impact of predictors through p-values and odds ratios (ORs). Specifically, for each type of cooperative perception error, the null hypothesis assumed no association between the number of errors and the likelihood of a violation. Associations are considered statistically significant when the p-value is below the predefined threshold of 0.05. The OR represents the ratio of the probability of a violation to the probability of it not occurring, and it quantifies the change in violation odds associated with each additional error occurrence.

\subsubsection{Experimental Results}

\begin{figure}[htbp]\small
	\centering
    \includegraphics[width=\linewidth]{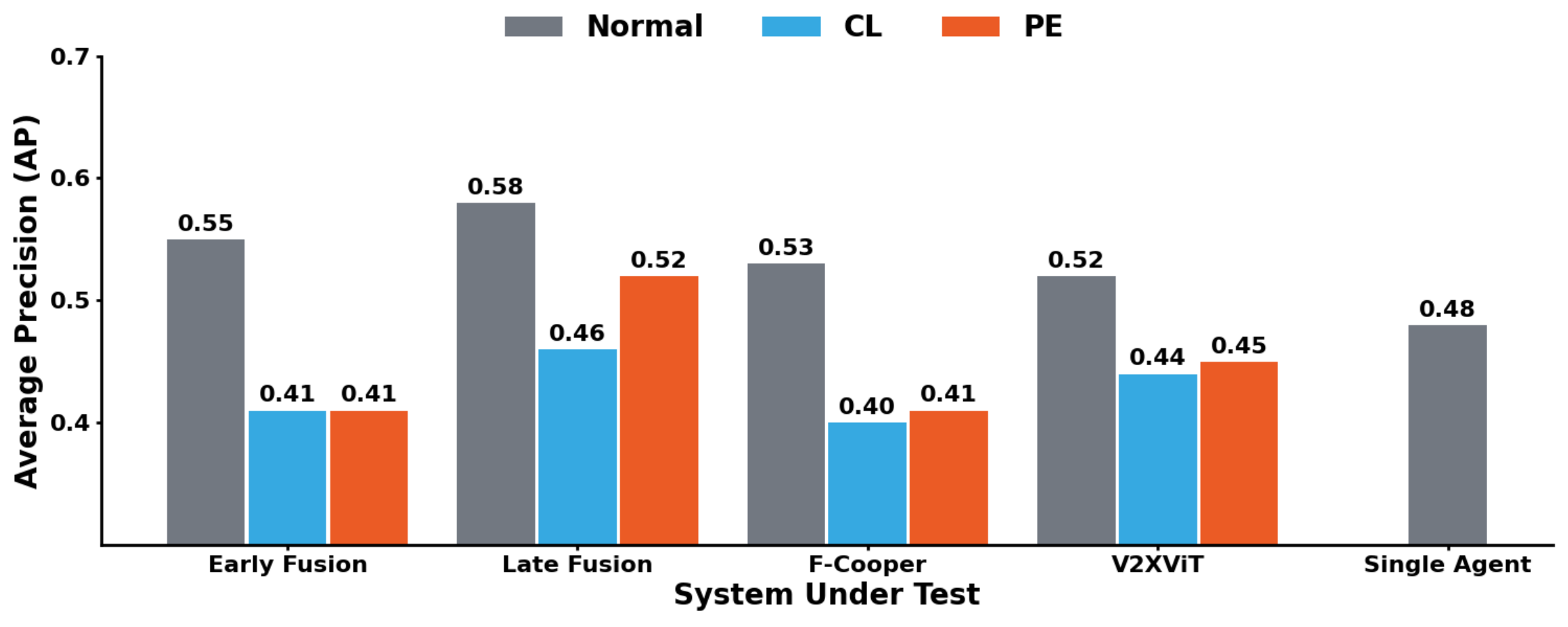}%
        \vspace{-5pt}
	\caption{Average precision of the system under test along the driving route.}
	\label{RQ3&4-single}
    \vspace{-25pt}
\end{figure}

Table~\ref{rq3-1} presents the driving performance of ADS equipped with different perception systems under normal communication conditions. The results indicate that all systems exhibit cooperative perception errors that lead to violations during online operation, with the V2XViT system achieving the highest overall driving score. Figure~\ref{RQ3&4-single} shows the AP of the system under test along the
driving route. As shown in Table~\ref{rq3-1} and Figure~\ref{RQ3&4-single}, under normal communication conditions, the cooperative perception system outperforms the single-agent perception system in terms of overall performance. Table~\ref{rq3-2} further provides an analysis of cooperative error patterns exhibited by the cooperative perception systems during route execution. The experimental results reveal that scenarios involving driving violations exhibit, on average, 15.4\% lower perception performance compared to non-violation scenarios, and the number of cooperative perception errors increases. To further investigate the underlying causes of these violations, the distribution of cooperative perception errors over the entire period from ADS initialization to the occurrence of a violation is provided on the supplementary website~\cite{website}. The results show that the frequency of cooperative perception errors increases as the time of the violation approaches, with particularly notable increases in CCME and CCLE. Furthermore, statistical analysis reveals that only CCLE is statistically significant ($p = 0.047$) under normal driving conditions, with an OR of 1.019, indicating a 1.9\% increase in the odds of violations for each additional CCLE error.

\vspace{-5pt}
\begin{center}
\begin{tcolorbox}[colback=gray!15,
                  colframe=black,
                  width=9cm,
                  arc=1mm, auto outer arc,
                  boxrule=0.5pt,size=title,opacityfill=0.1
                 ]
\textbf{Answer to RQ3:} Cooperative perception errors resulting from imperfect cooperative perception may lead to ADS violations. As the number of cooperative perception errors increases, the likelihood of violations also rises, with CCLE as a major contributor to driving violations.

\end{tcolorbox}
\end{center} 
\vspace{-10pt}

\begin{table}[htbp]
  \centering
  \vspace{-10pt}
  \caption{Driving performance of ADS equipped with different perception systems under normal and abnormal communication conditions.}
  \vspace{-5pt}
    \setlength{\tabcolsep}{4.8mm}
\begin{tabular}{lllll}
    \hline
    \multicolumn{1}{c|}{\textbf{Systems}} & \multicolumn{1}{c|}{\textbf{Type}} & \multicolumn{1}{c|}{\textbf{DS}} & \multicolumn{1}{c|}{\textbf{CR}} & \multicolumn{1}{c}{\textbf{RCR}} \bigstrut\\
    \hline
    \multicolumn{1}{c|}{\multirow{3}[6]{*}{\textbf{Early Fusion}}} & \multicolumn{1}{c|}{\textbf{Normal}} & \multicolumn{1}{c|}{\textbf{77.1 }} & \multicolumn{1}{c|}{\textbf{26.3 }} & \multicolumn{1}{c}{\textbf{79.4 }} \bigstrut\\
\cline{2-5}    \multicolumn{1}{c|}{} & \multicolumn{1}{c|}{\textbf{CL}} & \multicolumn{1}{c|}{72.9 } & \multicolumn{1}{c|}{31.6 } & \multicolumn{1}{c}{76.2 } \bigstrut\\
\cline{2-5}    \multicolumn{1}{c|}{} & \multicolumn{1}{c|}{\textbf{PE}} & \multicolumn{1}{c|}{64.2 } & \multicolumn{1}{c|}{42.1 } & \multicolumn{1}{c}{69.0 } \bigstrut\\
    \hline
    \multicolumn{1}{c|}{\multirow{3}[6]{*}{\textbf{Late Fusion}}} & \multicolumn{1}{c|}{\textbf{Normal}} & \multicolumn{1}{c|}{\textbf{78.9 }} & \multicolumn{1}{c|}{\textbf{26.3 }} & \multicolumn{1}{c}{\textbf{88.0 }} \bigstrut\\
\cline{2-5}    \multicolumn{1}{c|}{} & \multicolumn{1}{c|}{\textbf{CL}} & \multicolumn{1}{c|}{71.2 } & \multicolumn{1}{c|}{36.8 } & \multicolumn{1}{c}{83.0 } \bigstrut\\
\cline{2-5}    \multicolumn{1}{c|}{} & \multicolumn{1}{c|}{\textbf{PE}} & \multicolumn{1}{c|}{74.3 } & \multicolumn{1}{c|}{31.6 } & \multicolumn{1}{c}{83.7 } \bigstrut\\
    \hline
    \multicolumn{1}{c|}{\multirow{3}[6]{*}{\textbf{F-Cooper}}} & \multicolumn{1}{c|}{\textbf{Normal}} & \multicolumn{1}{c|}{\textbf{75.9 }} & \multicolumn{1}{c|}{\textbf{31.6 }} & \multicolumn{1}{c}{\textbf{81.7 }} \bigstrut\\
\cline{2-5}    \multicolumn{1}{c|}{} & \multicolumn{1}{c|}{\textbf{CL}} & \multicolumn{1}{c|}{71.2 } & \multicolumn{1}{c|}{31.6 } & \multicolumn{1}{c}{76.8 } \bigstrut\\
\cline{2-5}    \multicolumn{1}{c|}{} & \multicolumn{1}{c|}{\textbf{PE}} & \multicolumn{1}{c|}{62.5 } & \multicolumn{1}{c|}{42.1 } & \multicolumn{1}{c}{70.6 } \bigstrut\\
    \hline
    \multicolumn{1}{c|}{\multirow{3}[6]{*}{\textbf{V2XViT}}} & \multicolumn{1}{c|}{\textbf{Normal}} & \multicolumn{1}{c|}{\textbf{81.0 }} & \multicolumn{1}{c|}{\textbf{26.3 }} & \multicolumn{1}{c}{\textbf{87.9 }} \bigstrut\\
\cline{2-5}    \multicolumn{1}{c|}{} & \multicolumn{1}{c|}{\textbf{CL}} & \multicolumn{1}{c|}{65.1 } & \multicolumn{1}{c|}{42.1 } & \multicolumn{1}{c}{76.0 } \bigstrut\\
\cline{2-5}    \multicolumn{1}{c|}{} & \multicolumn{1}{c|}{\textbf{PE}} & \multicolumn{1}{c|}{77.2 } & \multicolumn{1}{c|}{26.3 } & \multicolumn{1}{c}{84.8 } \bigstrut\\
    \hline
    \multicolumn{1}{c|}{\textbf{Single Agent}} & \multicolumn{1}{c|}{\textbf{Normal}} & \multicolumn{1}{c|}{69.1} & \multicolumn{1}{c|}{36.8 } & \multicolumn{1}{c}{74.7 } \bigstrut\\
    \hline
    \multicolumn{5}{l}{\textbf{*Single-agent perception remains unaffected by CL and PE.}} \bigstrut[t]\\
    \end{tabular}%
  \vspace{-10pt}
  \label{rq3-1}%
\end{table}%

\subsection{RQ4. Cooperative perception systems are not robust against communication interference when running online.}

\subsubsection{Experimental Setup}
To investigate whether the driving performance of ADS equipped with the cooperative perception system is affected by common communication interferences, we introduce communication latency~(CL) and pose error~(PE) to the cooperative agents. Specifically, we utilize the test routes from RQ3, introduce two types of communication issues into these routes, and simulate them randomly within a predefined parameter range. 
In the experiment, we record the same information as in RQ3 (i.e., the number of cooperative perception errors and whether a violation occurred) for each test route. Based on this information, we then calculate the AP, DS, CR, and RCR. Similar to RQ3, to quantify the impact of different types of cooperative perception errors on driving violations under abnormal communication conditions, we use the multivariate logistic regression model~\cite{DBLP:journals/infsof/CerpaBKV10} for statistical analysis. Finally, to provide a more intuitive understanding, we select two representative collision cases attributable to communication latency and pose error for visualization.

\subsubsection{Experimental Results}

Table~\ref{rq3-1} presents the driving performance of ADS equipped with different cooperative perception systems under CL and PE. Compared to normal driving conditions, CL resulted in a 10.3\% decrease in driving score and a 6.3\% decrease in route completion rate, while the collision rate increased by 7.9\%. Similarly, under PE conditions, the driving score and route completion rate decreased by 11.2\% and 7.2\%, respectively, and the collision rate increased by 7.9\%. 
\begin{table}[htbp]
  \centering
  \vspace{-5pt}
  \caption{Online cooperative error pattern analysis of cooperative perception systems during route execution.}
    \vspace{-5pt}
    \setlength{\tabcolsep}{1.1mm}
    \renewcommand\arraystretch{0.95}
    \begin{tabular}{c|c|c|c|c|c|c|c|c}
    \hline
    \textbf{Systems} & \textbf{Type} & \textbf{LCME} & \textbf{LCLE} & \textbf{LADE} & \textbf{CCME} & \textbf{CCLE} & \textbf{CADE} & \textbf{AP} \bigstrut\\
    \hline
    \multirow{2}[4]{*}{\textbf{Early Fusion}} & \textbf{Vio.} & \textbf{2.8} & \textbf{4.0} & \textbf{22.6} & \textbf{78.6} & \textbf{29.8} & \textbf{15.0} & \textbf{0.50} \bigstrut\\
\cline{2-9}          & \textbf{No Vio.} & 2.3   & 2.3   & 14.4  & 51.9  & 26.6  & 6.5   & 0.56  \bigstrut\\
    \hline
    \multirow{2}[4]{*}{\textbf{Late Fusion}} & \textbf{Vio.} & \textbf{0.3} & \textbf{0.5} & \textbf{13.5} & \textbf{53.3} & \textbf{37.2} & \textbf{24.5} & \textbf{0.51} \bigstrut\\
\cline{2-9}          & \textbf{No Vio.} & 0.1   & 0.2   & 13.5  & 37.5  & 11.6  & 16.0  & 0.61  \bigstrut\\
    \hline
    \multirow{2}[4]{*}{\textbf{F-Cooper}} & \textbf{Vio.} & \textbf{1.3} & \textbf{3.2} & \textbf{27.8} & \textbf{54.3} & \textbf{55.8} & \textbf{38.8} & \textbf{0.44} \bigstrut\\
\cline{2-9}          & \textbf{No Vio.} & 0.7   & 0.8   & 12.1  & 51.4  & 27.1  & 25.8  & 0.56  \bigstrut\\
    \hline
    \multirow{2}[4]{*}{\textbf{V2XViT}} & \textbf{Vio.} & \textbf{1.6} & \textbf{10.8} & \textbf{18.4} & \textbf{95.4} & \textbf{63.4} & \textbf{14.0} & \textbf{0.47} \bigstrut\\
\cline{2-9}          & \textbf{No Vio.} & 0.6   & 3.9   & 13.1  & 63.9  & 20.1  & 10.1  & 0.54  \bigstrut\\
    \hline
    \end{tabular}%
    \vspace{-10pt}
  \label{rq3-2}%
\end{table}%
Figure~\ref{RQ3&4-single} shows the average precision of the system under CL and PE. As shown in Table~\ref{rq3-1} and Figure~\ref{RQ3&4-single}, under abnormal communication conditions, such as CL and PE, the perception accuracy and driving performance of certain cooperative perception systems may even fall below that of a single agent.
Table~\ref{rq4-1} shows an analysis of online error patterns of cooperative perception systems under abnormal communication conditions. From the table, we can conclude that the perception performance of the cooperative perception system is significantly reduced under the conditions of communication delay and pose error, and the number of most cooperative perception errors increases significantly. Communication latency tends to result in more cooperative missing errors, while pose errors are more likely to cause increased cooperative localization errors. 
Furthermore, our statistical analysis reveals that under the CL condition, three types of errors exhibit significant effects: LCLE ($p = 0.024$, OR = 1.137), CCME ($p = 0.023$, OR = 1.015), and CADE ($p = 0.016$, OR = 1.037). Under the PE condition, LCLE emerges as the most critical predictor ($p = 0.014$, OR = 1.169), corresponding to a 16.9\% increase in violation odds. These findings demonstrate that LCLE errors are the primary contributors to elevated violation rates under both CL and PE conditions. This can be attributed to the fact that abnormal communication conditions may transmit delayed or inconsistent information to the ego vehicle, thereby impairing its perception accuracy.

\begin{table*}[htbp]\scriptsize
  \centering
  \caption{Analysis of online error patterns of cooperative perception systems under abnormal communication conditions.}
  \vspace{-5pt}
    \setlength{\tabcolsep}{1.9mm}
    \begin{tabular}{c|c|c|c|c|c|c|c|c|c|c|c|c|c|c|c}
    \hline
    \multirow{2}[4]{*}{\textbf{Systems}} & \multirow{2}[4]{*}{\textbf{Type}} & \multicolumn{7}{c|}{\textbf{Communication Latency}}   & \multicolumn{7}{c}{\textbf{Pose Error}} \bigstrut\\
\cline{3-16}          &       & \textbf{LCME} & \textbf{LCLE} & \textbf{LADE} & \textbf{CCME} & \textbf{CCLE} & \textbf{CADE} & \textbf{AP} & \textbf{LCME} & \textbf{LCLE} & \textbf{LADE} & \textbf{CCME} & \textbf{CCLE} & \textbf{CADE} & \textbf{AP} \bigstrut\\
    \hline
    \multirow{2}[4]{*}{\textbf{Early Fusion}} & \textbf{Vio.} & \textbf{21.7}  & \textbf{12.2}  & \textbf{84.0} & \textbf{122.7} & \textbf{39.5} & \textbf{11.8} & \textbf{0.32} & \textbf{7.5} & \textbf{11.8} & \textbf{61.8} & \textbf{109.5} & \textbf{37.3} & \textbf{16.6} & \textbf{0.36} \bigstrut\\
\cline{2-16}          & \textbf{No Vio.} & 6.6   & 1.5   & 42.7  & 69.1  & 26.3  & 5.9   & 0.45  & 4.9   & 5.3   & 40.0  & 61.0  & 36.0  & 7.6   & 0.45 \bigstrut\\
    \hline
    \multirow{2}[4]{*}{\textbf{Late Fusion}} & \textbf{Vio.} & \textbf{0.1} & \textbf{1.3} & \textbf{47.8} & \textbf{92.0} & \textbf{30.8} & \textbf{34.4} & \textbf{0.38} & \textbf{0.0} & \textbf{3.1} & \textbf{29.1} & \textbf{67.7} & \textbf{57.1} & \textbf{19.6} & \textbf{0.45} \bigstrut\\
\cline{2-16}          & \textbf{No Vio.} & 0.0   & 0.0   & 24.5  & 27.3  & 11.0  & 11.2  & 0.52  & 0.0   & 0.8   & 27.0  & 38.8  & 18.3  & 15.2  & 0.57 \bigstrut\\
    \hline
    \multirow{2}[4]{*}{\textbf{F-Cooper}} & \textbf{Vio.} & \textbf{17.1} & \textbf{7.9} & \textbf{40.0} & \textbf{87.9} & 37.6  & \textbf{47.4} & \textbf{0.38} & \textbf{7.0} & \textbf{10.3} & \textbf{37.7} & \textbf{76.3} & \textbf{61.0} & \textbf{42.9} & \textbf{0.37} \bigstrut\\
\cline{2-16}          & \textbf{No Vio.} & 10.7  & 2.6   & 33.3  & 59.8  & \textbf{45.9} & 31.8  & 0.40  & 7.0   & 3.9   & 17.4  & 45.0  & 56.6  & 28.9  & 0.43 \bigstrut\\
    \hline
    \multirow{2}[4]{*}{\textbf{V2XViT}} & \textbf{Vio.} & \textbf{6.7} & \textbf{9.1} & \textbf{49.0} & \textbf{131.9} & \textbf{34.2} & \textbf{11.3} & \textbf{0.41} & \textbf{2.5} & \textbf{14.7} & \textbf{31.7} & \textbf{120.8} & \textbf{61.2} & \textbf{15.2} & \textbf{0.34} \bigstrut\\
\cline{2-16}          & \textbf{No Vio.} & 1.6   & 4.7   & 36.9  & 72.7  & 17.3  & 8.5   & 0.46  & 2.5   & 4.0   & 21.5  & 62.1  & 18.4  & 9.7   & 0.49 \bigstrut\\
    \hline
    \end{tabular}%
    \vspace{-10pt}
  \label{rq4-1}%
\end{table*}%

\begin{figure*}[htbp]
    \centering
    \subfigure[$time_0$]{
    \includegraphics[width=1.3in]{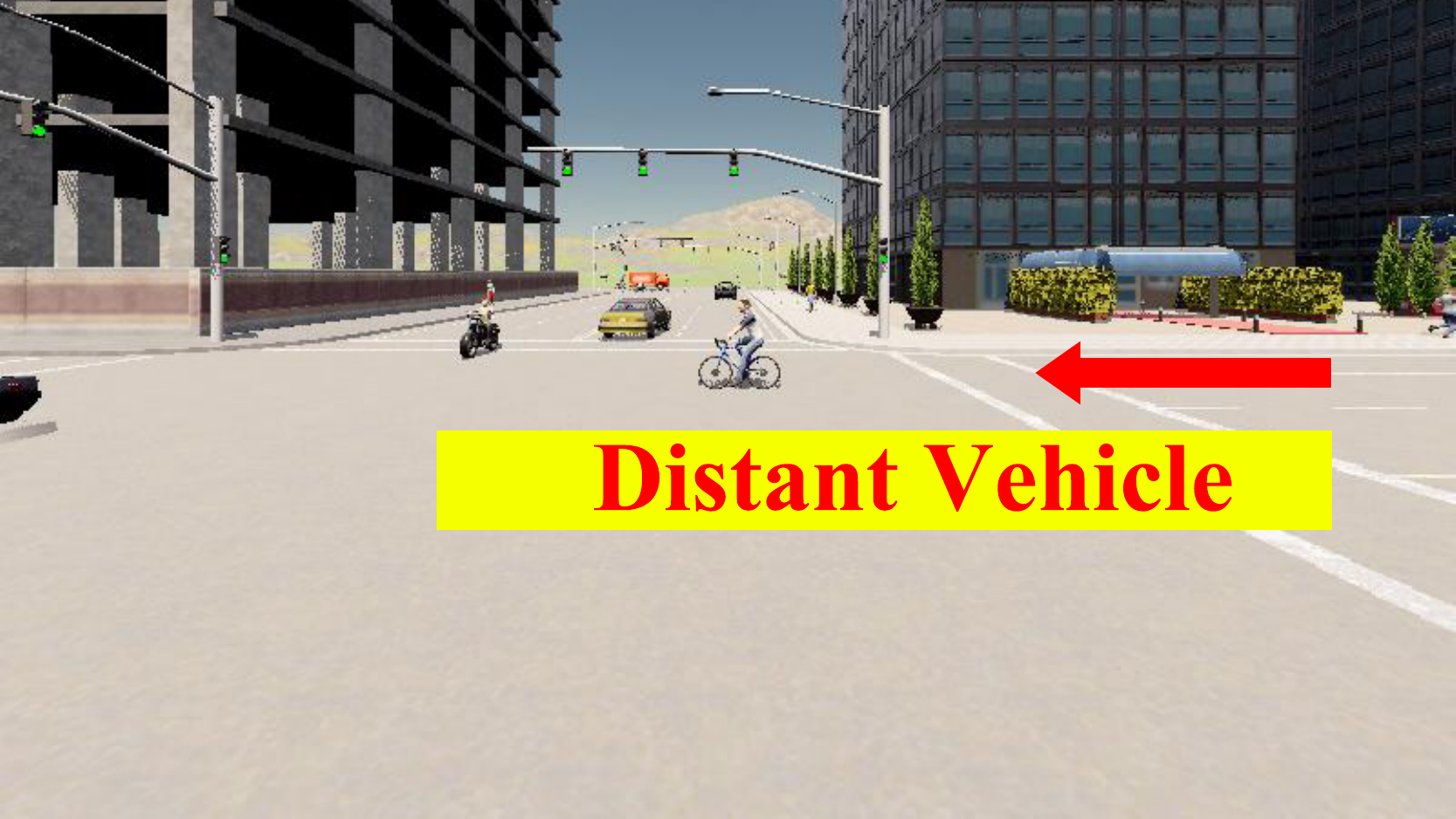}
    }
    \subfigure[$time_1$(Normal)]{
	\includegraphics[width=1.3in]{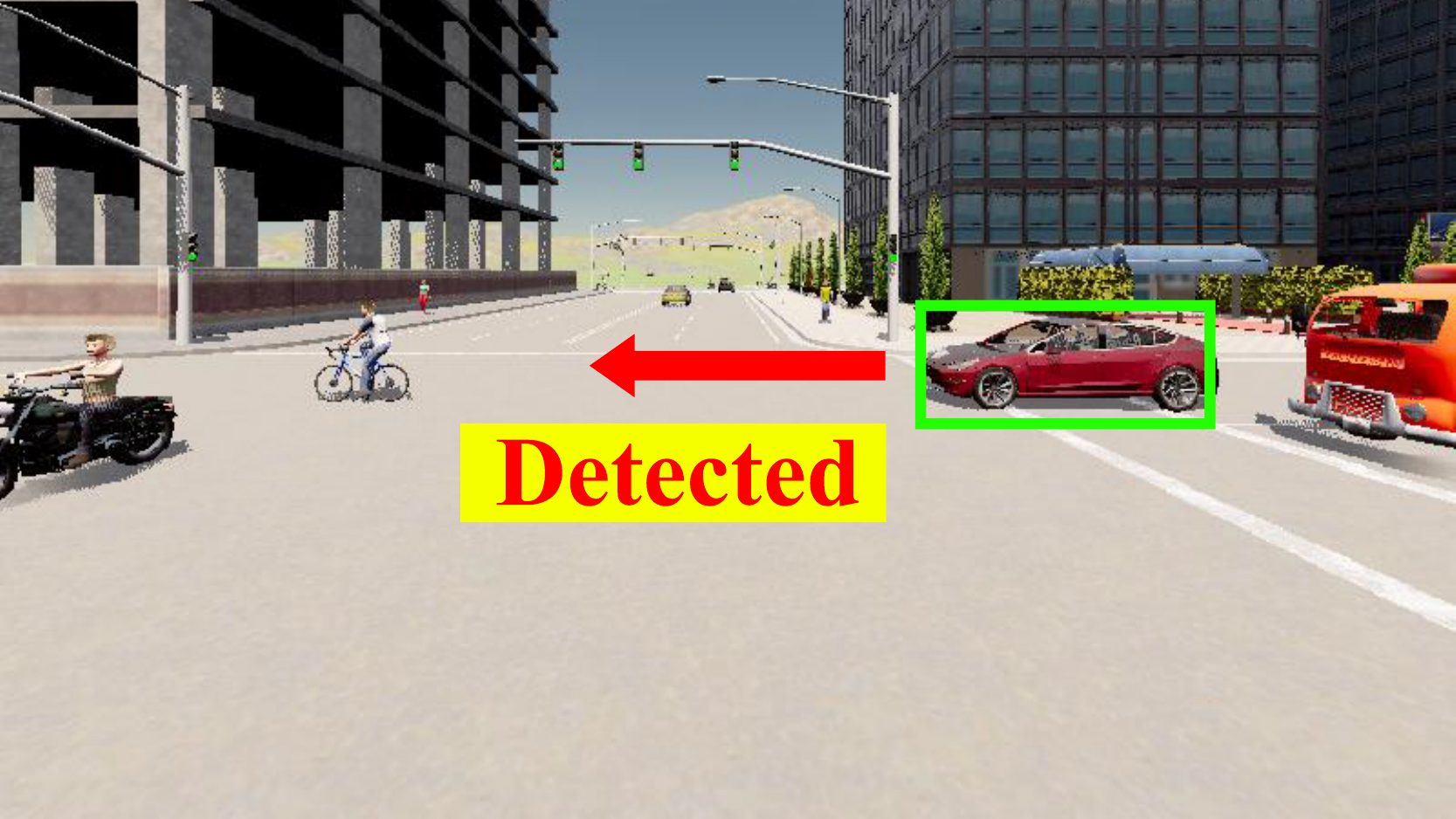}
    }
    \subfigure[$time_2$(Normal)]{
	\includegraphics[width=1.3in]{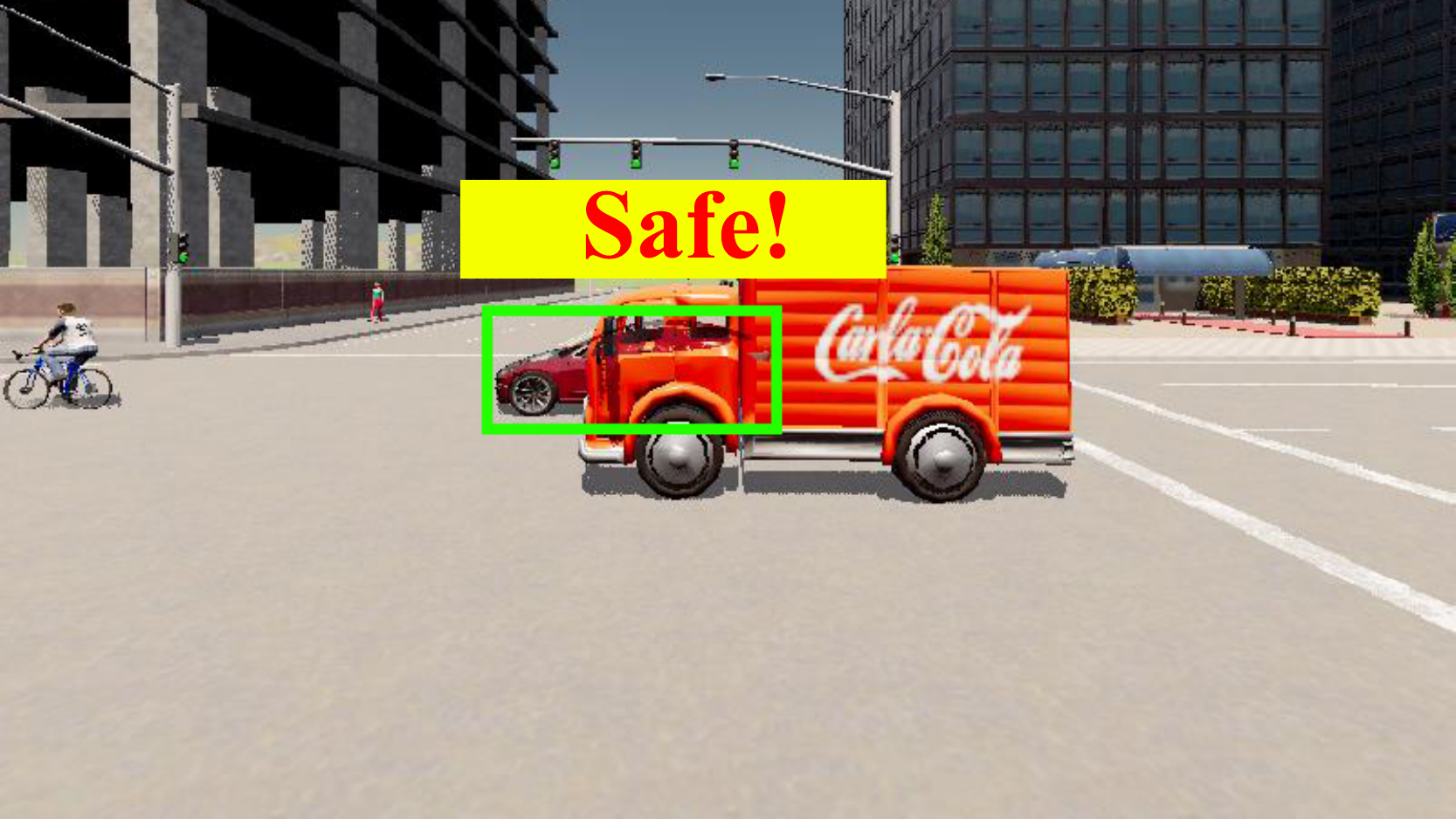}
    }
    \subfigure[$time_1$(CL)]{
	\includegraphics[width=1.3in]{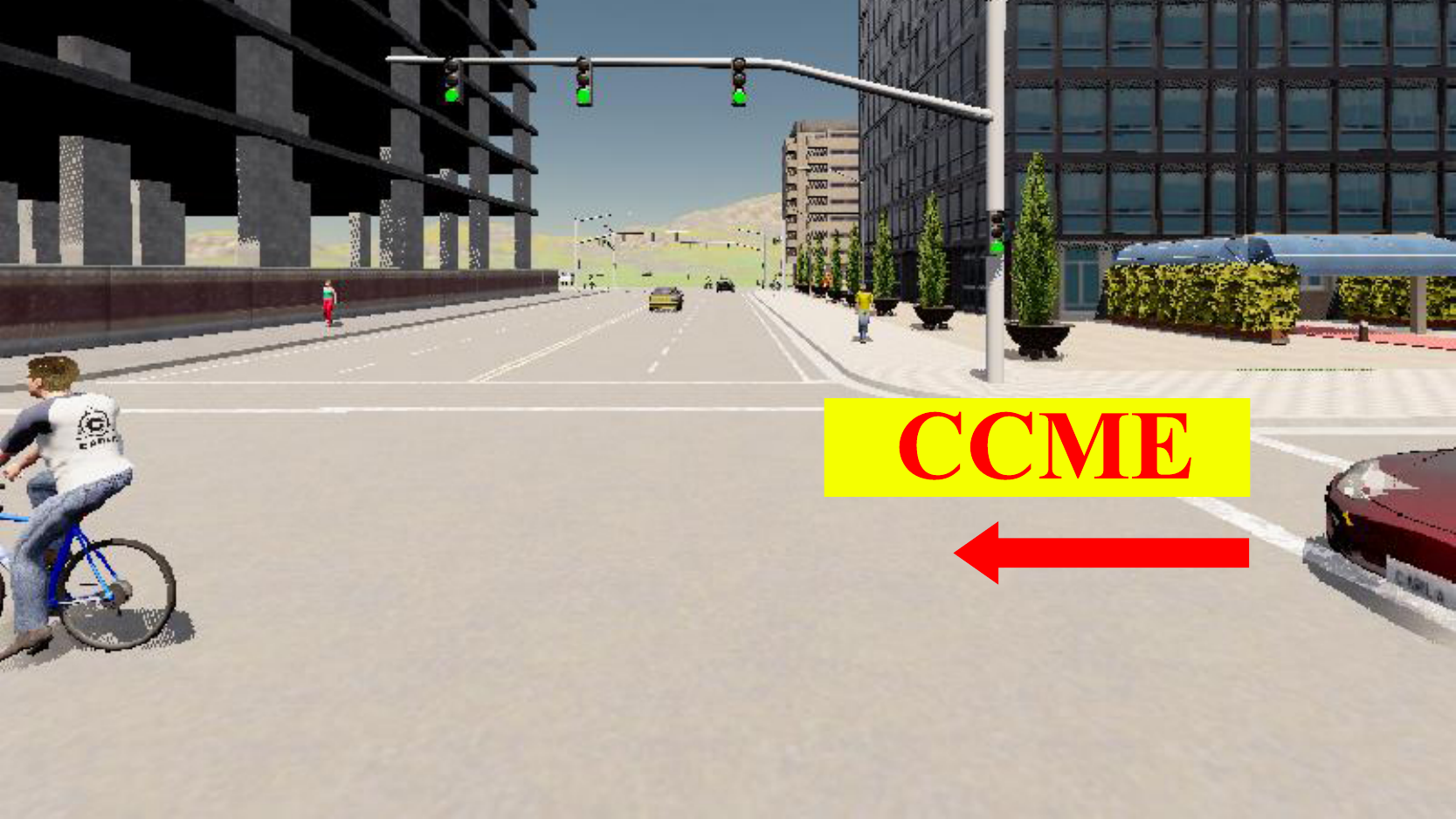}
    }
    \subfigure[$time_2$(CL)]{
	\includegraphics[width=1.3in]{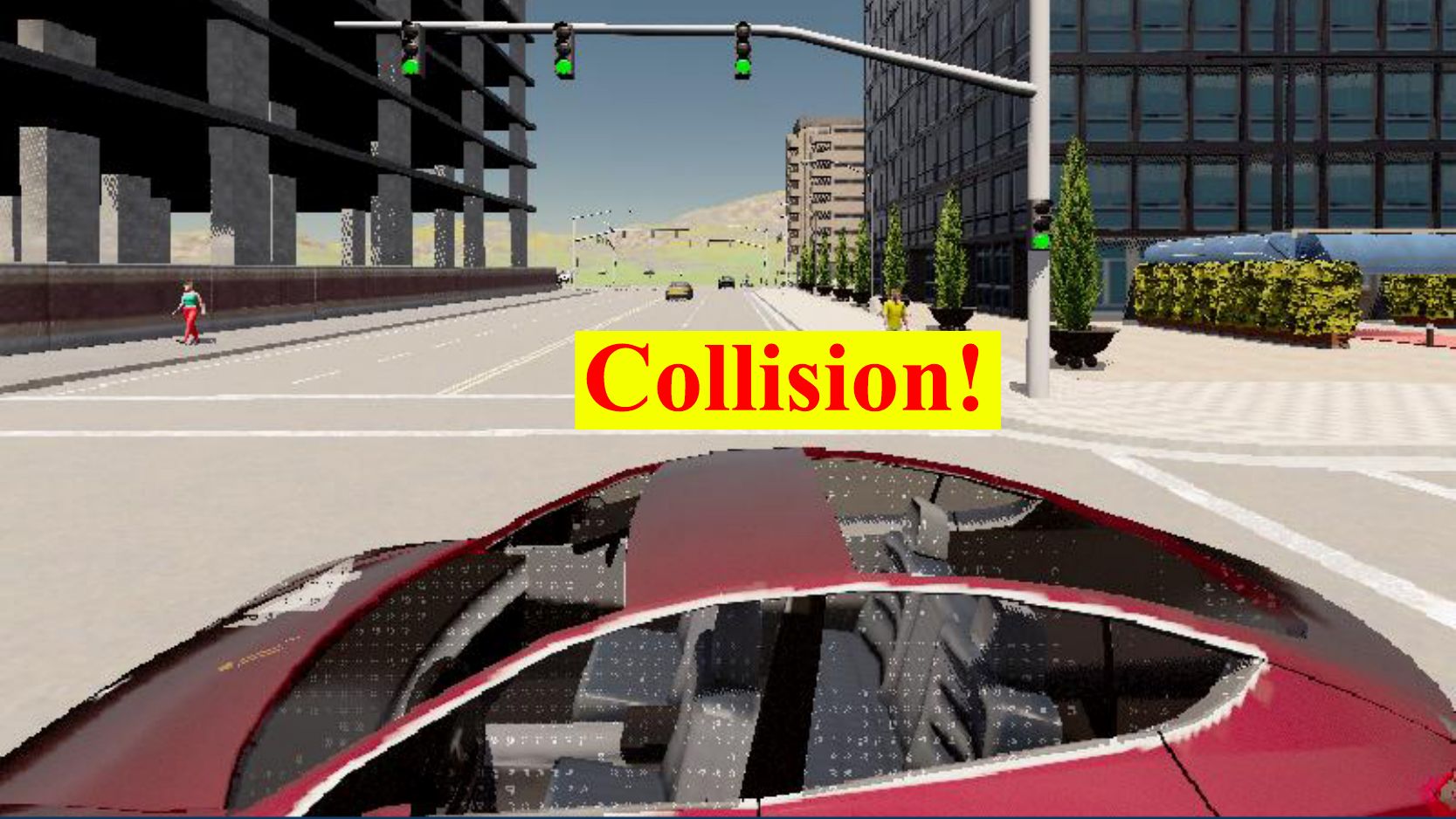}
    }
    \subfigure[$time_0$]{
    \includegraphics[width=1.3in]{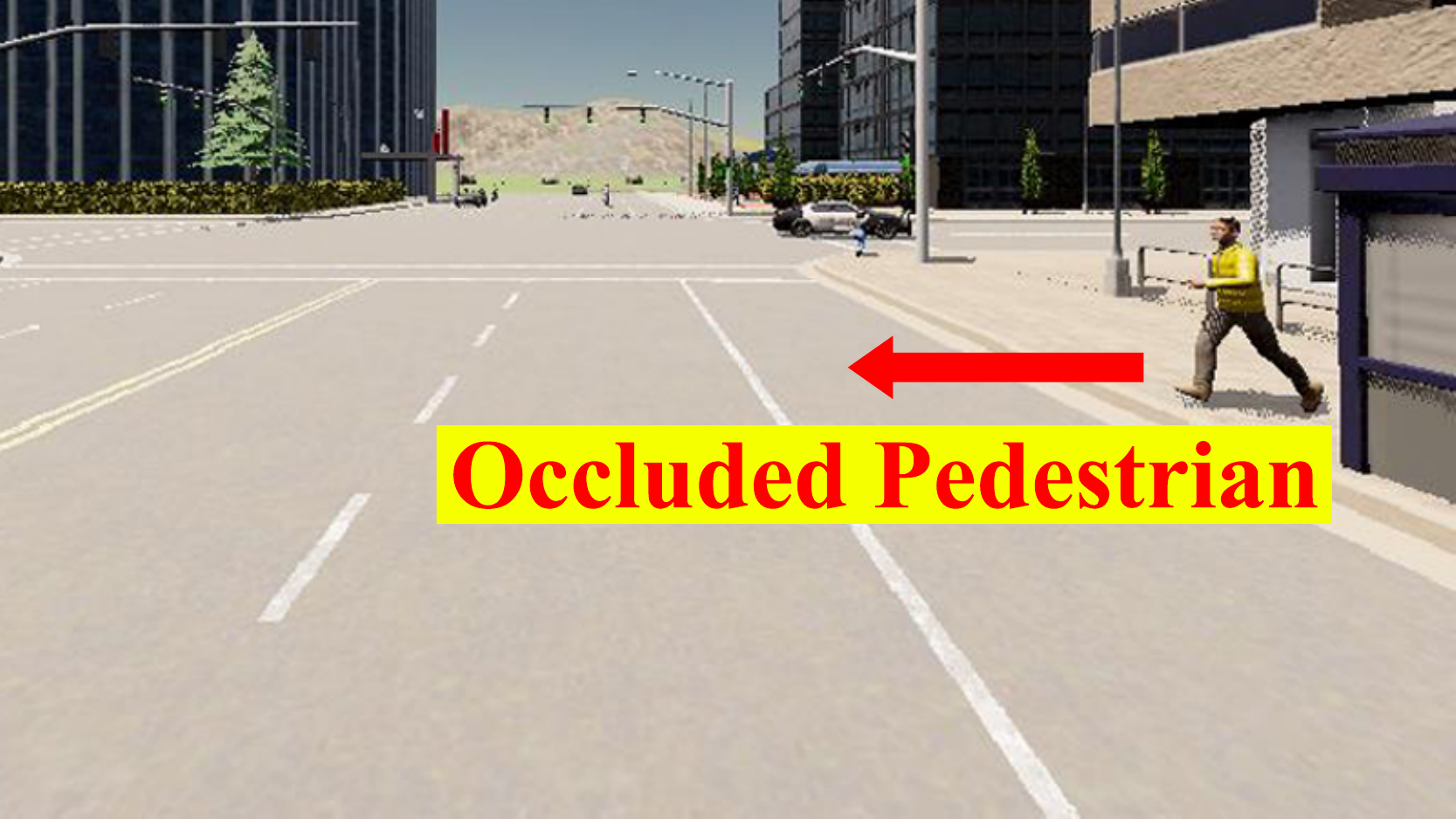}
    }
    \subfigure[$time_1$(Normal)]{
	\includegraphics[width=1.3in]{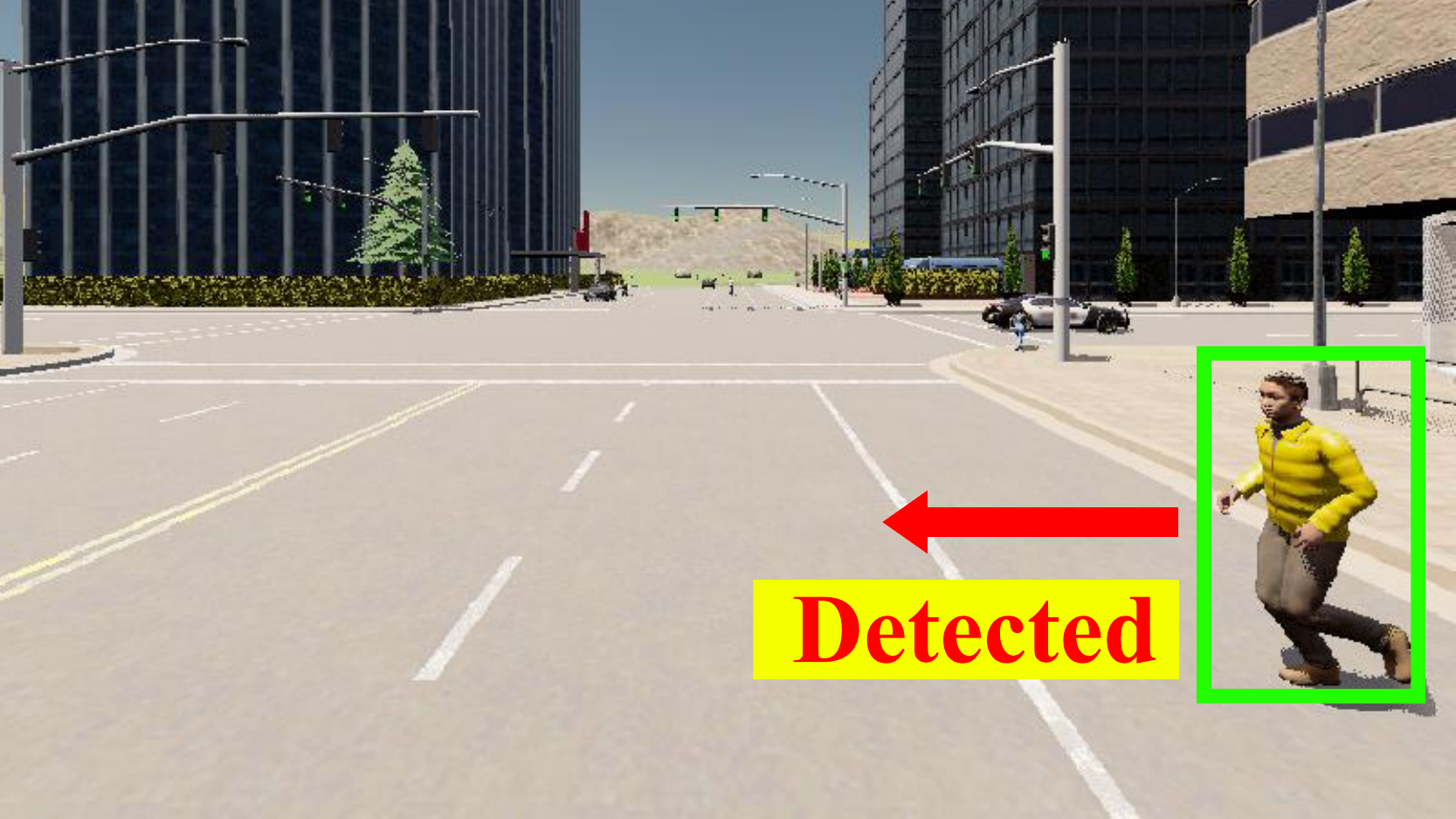}
    }
    \subfigure[$time_1$(Normal)]{
	\includegraphics[width=1.3in]{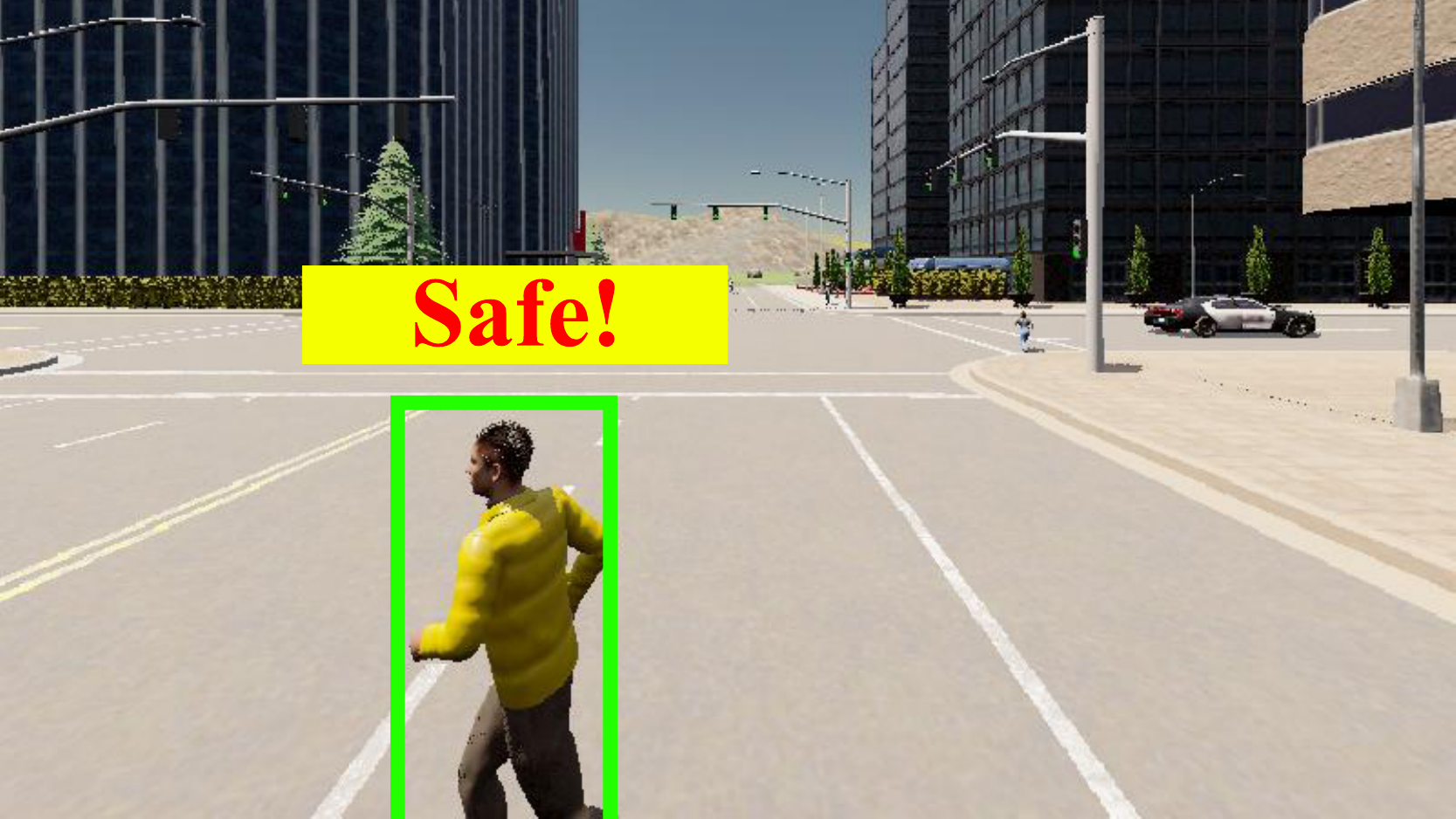}
    }
    \subfigure[$time_2$(PE)]{
	\includegraphics[width=1.3in]{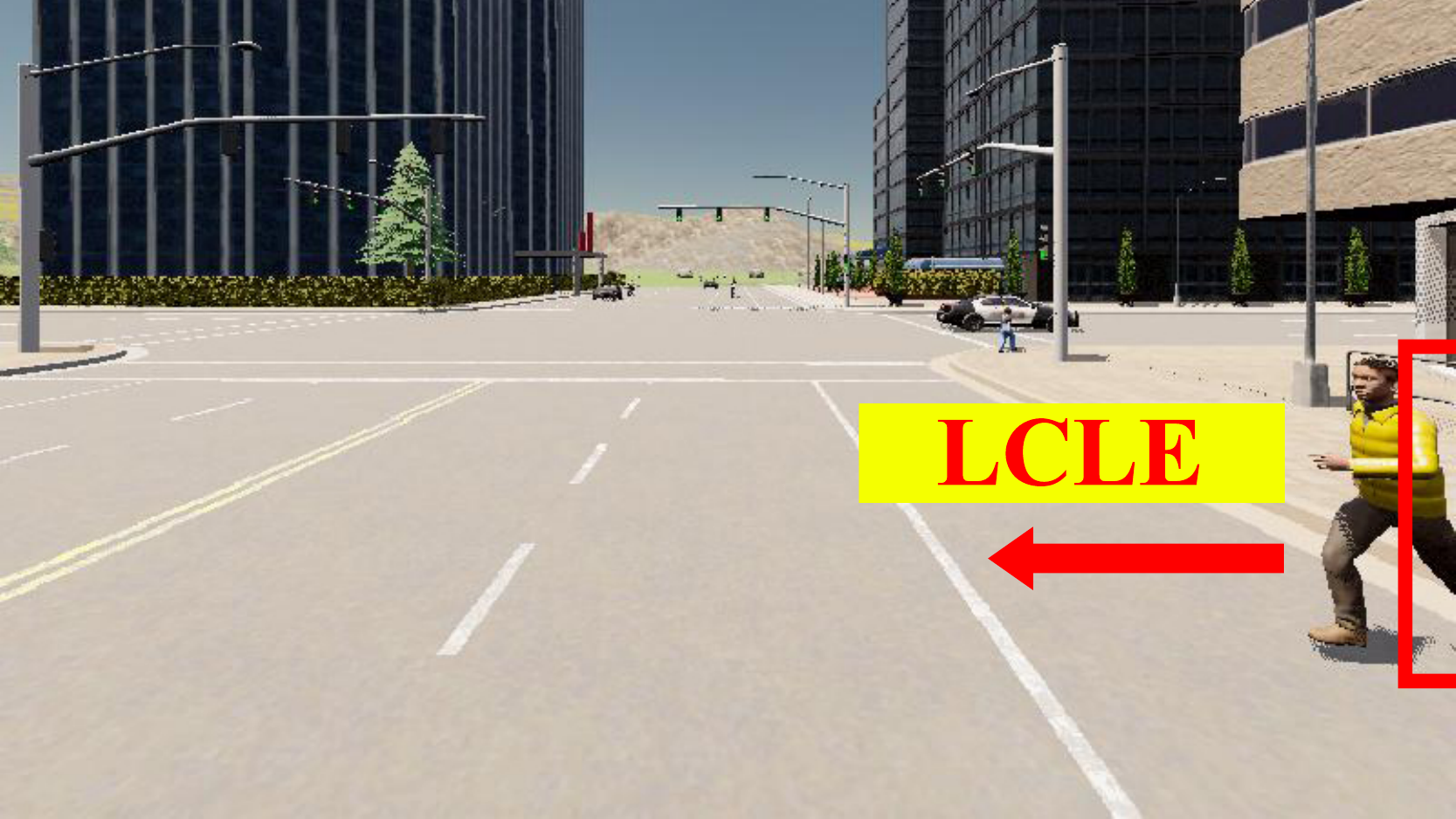}
    }
    \subfigure[$time_2$(PE)]{
	\includegraphics[width=1.3in]{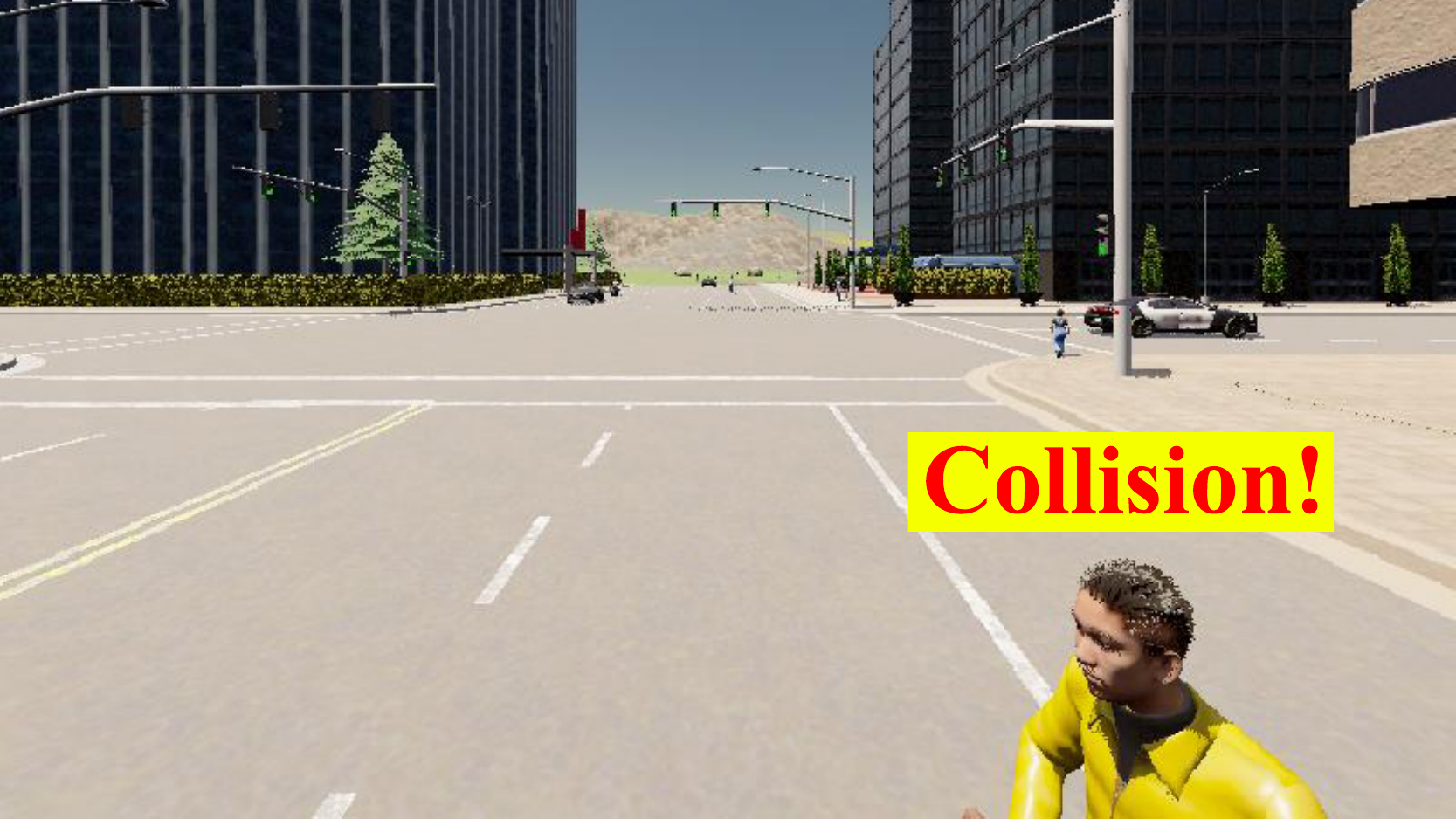}
    }
    \vspace{-10pt}
    \caption{Visualization samples showing representative collision cases attributable to communication latency and pose error. }
    \label{rq4-3}
    \vspace{-20pt}
\end{figure*}

Figure~\ref{rq4-3} presents visualization samples illustrating representative collision cases caused by communication latency and pose errors. $time_0$ represents the initial timestamp of the scenario, while $time1$ and $time_2$ denote timestamps corresponding to different moments during the scenario's execution under varying communication conditions. The first scenario~(top row of Figure~\ref{rq4-3}) involves a vehicle approaching from a distance and crossing an intersection. Under normal communication conditions, the ADS equipped with a cooperative perception system successfully detects the vehicle and decelerates accordingly. However, under communication delay, the system experiences a CCME, causing the ADS to fail to slow down and subsequently collide with the vehicle. The second scenario~(second row of Figure~\ref{rq4-3}) involves an occluded pedestrian crossing the road. Under normal communication conditions, the ADS equipped with a cooperative perception system detects the pedestrian in advance and stops to yield. However, in scenarios involving pose errors, the system encounters a LCLE and misjudges the pedestrian’s position, causing the ADS to fail to decelerate in time and ultimately resulting in a collision. These two representative scenarios highlight the lack of robustness of V2X cooperative perception systems under abnormal communication conditions.

\vspace{-10pt}
\begin{center}
\begin{tcolorbox}[colback=gray!15,
                  colframe=black,
                  width=9cm,
                  arc=1mm, auto outer arc,
                  boxrule=0.5pt,size=title,opacityfill=0.1
                 ]
\textbf{Answer to RQ4:} Abnormal communication conditions (such as communication latency and pose error) can undermine the performance of the cooperative perception system and significantly increase the frequency of ADS violations. Under CL conditions, LCLE, CCME, and CADE are the primary contributors to driving violations, whereas under PE conditions, LCLE emerges as the dominant factor.
\end{tcolorbox}
\end{center} 
\vspace{-10pt}

\section{Discussion}

Based on the findings from RQ1–RQ4, existing cooperative perception systems require further improvement for practical use. First, such systems must operate in heterogeneous environments, where sensor configurations often prioritize LiDAR over cameras due to LiDAR’s superior depth perception capabilities (as point clouds inherently contain spatial information). Multimodal sensor fusion designs must balance cost-effectiveness and communication efficiency. In addition, V2V cooperation generally outperforms V2I under intermediate and late fusion, likely because infrastructure-mounted sensors capture a bird's-eye view that is harder to fuse with ego vehicle features. Conversely, V2I cooperation shows superior performance in early fusion, potentially due to the infrastructure being equipped with higher resolution sensors. Considering that in the real world, the ego vehicle continuously interacts with diverse heterogeneous agents, cooperative perception systems should adapt to such conditions to ensure robust and reliable cooperative perception.

When an ADS equipped with a cooperative perception system operates online, communication delays and pose errors may further exacerbate ADS violations. In the real world, these two types of errors are prevalent. Even well-calibrated and synchronized sensors may still be inaccurate due to changes in the external environment. In order to deploy reliable cooperative perception systems, developers must carefully handle agent synchronization and calibration issues and design cooperative perception systems that can remain reliable in the face of abnormal communications.

We systematically summarize the error patterns of cooperative perception systems and find that a substantial number of cooperative perception errors occur during both offline and online operation. The data-driven nature makes it challenging to train a robust cooperative perception system that satisfies safety and reliability requirements under all conditions. Consequently, cooperative perception system development must prioritize cooperative perception error mitigation in design and training processes.

\noindent\textbf{Future Directions.} 
Based on these insights, we summarize the following future directions:

\vspace{-5pt}
\begin{itemize}

\item Cooperative perception systems should adapt to heterogeneous environments, where variations in varying sensor types and cooperative agent configurations could lead to inconsistent performance. Therefore, \textbf{the ongoing development of systems adapting to heterogeneous environments, coupled with the establishment of appropriate testing and optimization platforms, is essential.}

\item There is an urgent need for techniques to continuously enhance the robustness and reliability of V2X cooperative perception systems. Given that V2X involves multiple critical components, \textbf{robustness assurance should encompass both offline and online aspects, including offline fault localization and repair, as well as monitoring under abnormal communications and perception during system online operation.}

\item Cooperative perception systems should not perform worse than ego vehicle perception. However, our study systematically reveals the presence of cooperative perception error patterns, which can degrade system perception and even lead to driving accidents. \textbf{Designing efficient automated testing methodologies based on these patterns to detect and mitigate such cooperation errors represents a promising direction for future research.}

\end{itemize}

\noindent\textbf{Threats to Validity.} 
In terms of \textit{construct validity}, a potential threat arises from the classification of cooperative perception errors. Although this paper identifies six error patterns from the perspective of how cooperation influences the ego vehicle’s perception in the cooperative object detection task, alternative definitions of error patterns may exist under different viewpoints or cooperative perception tasks. Nevertheless, within the scope of this study, our classification is regarded as a complete and representative characterization of cooperative perception errors.
In terms of \textit{internal validity}, one of the primary threats is to conduct evaluations in a simulated environment. Simulated driving scenarios may differ both visually and physically from real-world environments due to the inherent gap between simulation-based testing and testing in the physical world. To alleviate this threat, we use the popular high-fidelity CARLA simulator to collect offline data or simulate online simulation scenarios, and we conduct multiple repeated experiments to ensure the reproducibility of our results. 
In terms of \textit{external validity}, a potential threat is that our results may not generalize to other cooperative perception systems, datasets, or cooperation modes. To mitigate this threat, we try our best to collect a diverse set of cooperative perception systems with varying system structures and fusion schemes, and adopt two representative cooperation modes (V2V and V2I). We then evaluate them using widely adopted datasets with rich scenarios and diverse test routes across a variety of testing settings to ensure comprehensive assessment. Another potential threat lies in our task choice, as this study mainly focuses on cooperative object detection. Nevertheless, as one of the most representative tasks in current research on cooperative perception, cooperative object detection provides the most direct and critical input for downstream autonomous driving tasks, such as object tracking and motion planning. In future work, we plan to extend our classification approach to other cooperative perception tasks.

\section{Related Works}

\noindent\textbf{V2X Cooperative Perception.} Cooperative perception systems facilitate the exchange of perception data among agents, thereby enabling each agent to achieve more comprehensive and accurate situational awareness. In this context, early fusion techniques involve the transmission of raw sensor data~\cite{DBLP:conf/icdcs/ChenTYF19}, whereas late fusion techniques transmit processed perception outputs~\cite{xu2022opv2v}. To balance computational performance and bandwidth efficiency, recent studies have investigated intermediate feature transmission~\cite{DBLP:journals/itsm/HanZLJLL23}. The first intermediate cooperative framework F-Cooper~\cite{DBLP:conf/edge/ChenMTGYF19} balances bandwidth constraints and detection accuracy by using max pooling for V2V voxel feature fusion. To consider the potential relationships between multiple agents, several graph-based fusion approaches have been proposed~\cite{DBLP:conf/eccv/WangMLYZU20,DBLP:conf/nips/LiRWCFZ21,DBLP:journals/ral/ZhouXZL22}. For instance, V2VNet~\cite{DBLP:conf/eccv/WangMLYZU20} leverages a spatial-aware graph neural network to model the communication among agents.
In addition to graph learning, attention mechanisms have emerged as a powerful tool for exploring feature relationships~\cite{DBLP:conf/nips/VaswaniSPUJGKP17,DBLP:conf/corl/XuTXSZM22,DBLP:conf/eccv/XuXTXYM22,tan2024dynamic,DBLP:journals/tiv/LinTDZZC24}. V2X-ViT~\cite{DBLP:conf/eccv/XuXTXYM22} proposes a multi-scale window attention module to capture long-range spatial interaction on high-resolution detection. Recent studies have also focused on developing cooperative perception systems that remain robust under communication interruptions~\cite{DBLP:journals/tiv/RenLWDWCZ24} and lossy transmission conditions~\cite{DBLP:journals/tiv/LiXLMCMY23}. However, current literature lacks investigation into how heterogeneous sensor configurations and varying cooperative agents affect the performance of cooperative perception systems, nor have such systems been empirically tested through online deployment.

\noindent\textbf{Autonomous Driving Perception System Testing.} 
Safety is a critical priority in the development of ADSs, and various approaches, such as data-driven~\cite{DBLP:conf/kbse/00020TFSSGL0024,DBLP:conf/sigsoft/GambiHF19,DBLP:conf/issta/Zhang023} and search-based~\cite{DBLP:conf/icse/HuaiCANLWCG23,DBLP:conf/issta/HildebrandtSE23,DBLP:journals/tse/ZhongKR23,DBLP:conf/aaai/TianHWG0ZL0025,DBLP:journals/jss/GuoFCC24} methods, have been proposed to test their safety. The perception system in an ADS serves as a key component in real-world applications, and rigorous testing is required to ensure the reliability of the overall system~\cite{DBLP:journals/spm/LiI20}. Various studies have focused on testing single-agent perception systems across various driving tasks~\cite{wang2020metamorphic,DBLP:conf/issta/GuoF022,christian2023generating}. To evaluate camera-based perception systems, a representative study~\cite{wang2020metamorphic} generates data by incorporating object instances into background images.
Several studies also test LiDAR-based perception systems~\cite{DBLP:conf/issta/GuoF022,christian2023generating}. Christian et al.~\cite{christian2023generating} propose testing LiDAR systems by introducing semantic mutations to real-world data for the semantic segmentation task. Recently, due to the rapid development of V2X communication technology, some researchers have focused on the testing of cooperative perception systems~\cite{DBLP:conf/issta/GuoG00LGSF24,DBLP:conf/issre/LiZTZ024,DBLP:conf/vtc/HawladerRF24}. Li et al.~\cite{DBLP:conf/issre/LiZTZ024} focus on investigating the impact of variations and disturbances in the communication range on safety and potential hazards. Guo et al. pioneered CooTest ~\cite{DBLP:conf/issta/GuoG00LGSF24}, a framework for evaluating cooperative perception robustness under extreme conditions such as communication interference and adverse weather.
CooTest~\cite{DBLP:conf/issta/GuoG00LGSF24}, the most related work, lacks a comprehensive analysis of cooperative perception errors, is limited to LiDAR-based V2V cooperation, and operates offline. In contrast, this paper analyzes error patterns in depth, considers heterogeneous sensors (LiDAR and camera) and diverse cooperative agents (V2V and V2I), and is the first to evaluate cooperative perception performance in the online deployment.

\section{Conclusion}

This paper presents an empirical study of V2X cooperative perception aimed at understanding the limitations and potential risks of cooperative perception systems in real-world applications. To achieve this, we conduct a comprehensive analysis of error patterns and perform large-scale performance evaluations for multiple critical components of cooperative perception systems.
First, we evaluate the performance differences of cooperative perception systems across diverse sensor types and cooperative agent configurations in heterogeneous environments. Additionally, we evaluate the cooperative perception systems online across 32 test routes, assessing their performance under both normal and abnormal communication conditions, including communication latency and pose errors.
Our findings reveal potential vulnerabilities in key components of cooperative perception systems. Finally, we outline several promising directions for future research to develop more reliable and robust V2X cooperative perception systems.

\balance
\bibliographystyle{IEEEtran}
\bibliography{mybib}

\end{document}